\documentclass{article} 
\usepackage[preprint]{colm2026_conference}

\usepackage{microtype}
\usepackage{xcolor}
\usepackage{hyperref}
\usepackage{url}
\usepackage{booktabs}
\usepackage{amsmath, amsfonts, amssymb}
\usepackage{graphicx}
\usepackage{multirow}
\usepackage{enumitem}
\usepackage{algorithm}
\usepackage{algorithmic}
\usepackage{subcaption}
\usepackage{xspace}

\usepackage{lineno}

\definecolor{darkblue}{rgb}{0, 0, 0.5}
\hypersetup{colorlinks=true, citecolor=darkblue, linkcolor=darkblue, urlcolor=darkblue}

\newcommand{\system}{\textsc{AlphaLab}\xspace}

\title{\system: Autonomous Multi-Agent Research \\ Across Optimization Domains with Frontier LLMs}

\author{
  Brendan R.\ Hogan\thanks{\scriptsize Corresponding author: \texttt{brendan.rappazzo@morganstanley.com}. Other authors: \texttt{firstname.lastname@morganstanley.com}.} \And Xiwen Chen \And James T.\ Wilson \And Kashif Rasul
  \AND
  Adel Boyarsky \And Thomas Kamei \And Anderson Schneider \And Yuriy Nevmyvaka
  \AND
  \mdseries Morgan Stanley
}

\begin{document}

\ifcolmsubmission
\linenumbers
\fi

\maketitle

\begin{abstract}
We present \system, an autonomous research harness that leverages frontier LLM agentic capabilities to automate the full experimental cycle in quantitative, computation-intensive domains.
Given only a dataset and a natural-language objective, \system proceeds through three phases without human intervention: (1)~it adapts to the domain and explores the data, writing analysis code and producing a research report; (2)~it constructs and adversarially validates its own evaluation framework; and (3)~it runs large-scale GPU experiments via a Strategist/Worker loop, accumulating domain knowledge in a persistent playbook that functions as a form of online prompt optimization.
All domain-specific behavior is factored into adapters generated by the model itself, so the same pipeline handles qualitatively different tasks without modification.
We evaluate \system with two frontier LLMs (GPT-5.2 and Claude Opus~4.6) on three domains: CUDA kernel optimization, where it writes GPU kernels that run $4.4\times$ faster than \texttt{torch.compile} on average (up to $91\times$); LLM pretraining, where the full system achieves 22\% lower validation loss than a single-shot baseline using the same model; and traffic forecasting, where it beats standard baselines by 23--25\% after researching and implementing published model families from the literature.
The two models discover qualitatively different solutions in every domain (neither dominates uniformly), suggesting that multi-model campaigns provide complementary search coverage.
We additionally report results on financial time series forecasting in the appendix, and release all code at \url{https://brendanhogan.github.io/alphalab-paper/}.
\end{abstract}

\section{Introduction}
\label{sec:intro}

Frontier LLMs have dramatically improved in agentic and software development ability since December 2025 .
These models are genuinely \emph{agentic}: given access to a terminal, a file system, and the internet, they can operate autonomously over extended periods, writing and debugging code, searching the web, installing dependencies, and iterating on their own output until a task is complete.
On SWE-bench Verified, solve rates among top agentic systems rose from roughly 30\% to over 70\% in under twelve months \citep{jimenez2024swebench}; the duration of tasks that models can sustain autonomously has grown from minutes to hours.

\begin{figure}[t]
\centering
\includegraphics[width=0.7\textwidth]{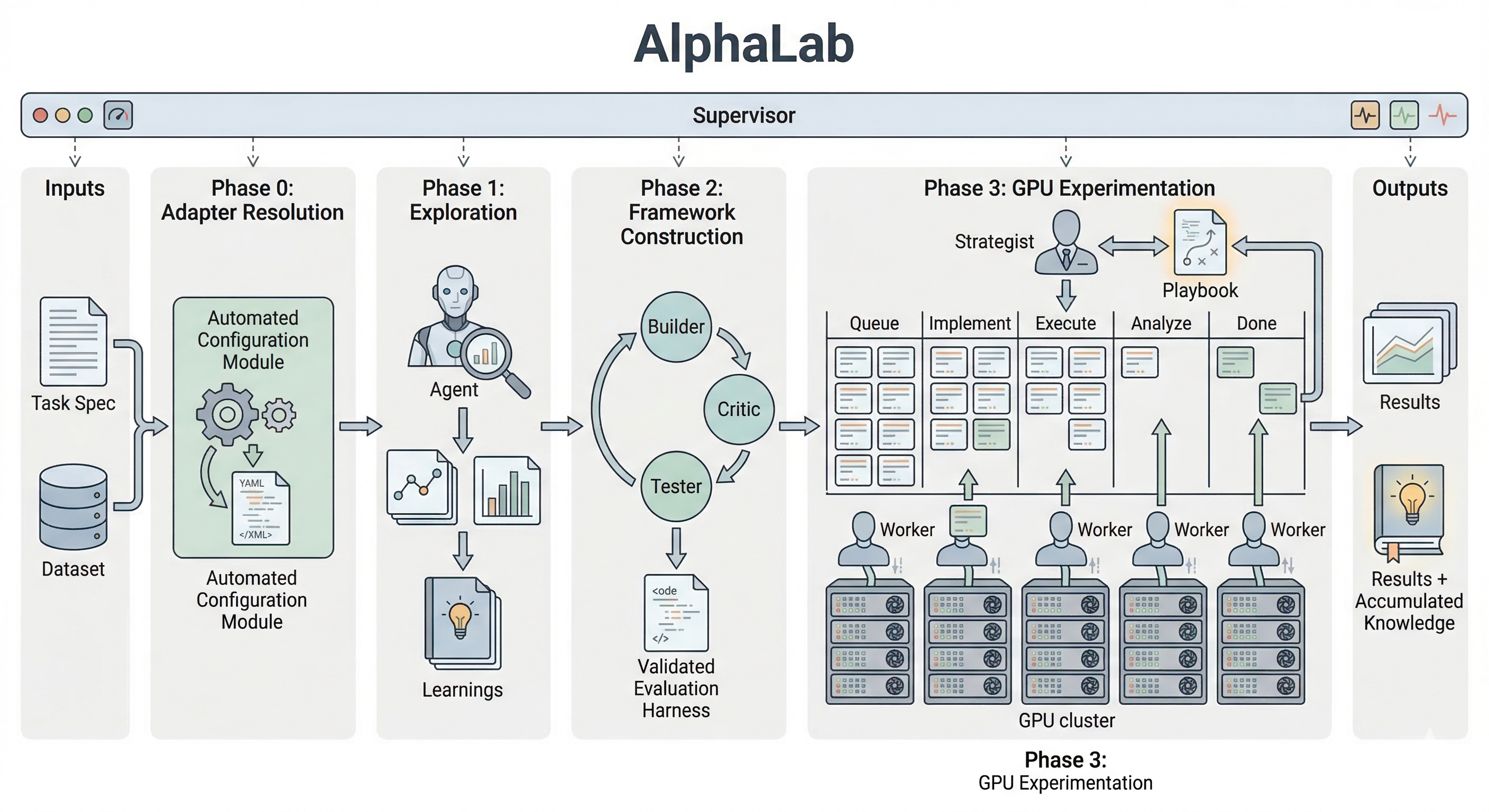}
\caption{\textbf{\system pipeline overview.}  Given a dataset and objective, the system writes all its own code across four phases.  \textbf{Phase~0:} configures domain-specific prompts/metrics.  \textbf{Phase~1:} explores the dataset and researches prior work via web search.  \textbf{Phase~2:} adversarial Builder/Critic/Tester loop constructs the evaluation framework.  \textbf{Phase~3:} Strategist proposes experiments, Workers implement them, and jobs run on a GPU cluster via Slurm; a persistent Playbook evolves with each experiment.  A Supervisor monitors health.  The human can interact with the Strategist to guide search; we run fully autonomously in this paper for fair comparison.}
\label{fig:pipeline}
\end{figure}

This shift has a compounding consequence.
The \emph{harnesses} that make these models effective (prompts, tool configurations, scaffolding code, evaluation pipelines) are themselves just software that the models are now good enough to write.
Products like Claude Code \citep{claudecode2025} and Codex \citep{codex2025} already demonstrate this: a frontier model builds its own tooling, evaluates results, and revises the tooling in a loop of recursive self-improvement.
For a growing class of problems, the answer is not fine-tuning but \emph{harness engineering}: building the right scaffolding and letting the model refine it \citep{poetiq2025arcagi,chollet2025arcagi2}.
If this works for software engineering, an obvious question is: what about scientific research?

Automating science occupies a unique position in the landscape of AI ambitions \citep{lu2024aiscientist,lu2025aiscientistv2,jiang2025aide,schmidgall2025agentlab}.
The scientific method (forming hypotheses, designing experiments, updating beliefs from evidence) is arguably what most distinguishes human intelligence from other forms of cognition.
A system that can do this fully end-to-end (autonomously identifying important problems, designing experiments, and producing genuine insight) would represent something close to artificial general intelligence.
In our view, current models are not there yet.
But even in the regime where AI serves as a very sophisticated tool rather than an autonomous scientist, the practical impact is enormous: for any domain with objective, easy-to-evaluate metrics (training loss, kernel speedup, forecast accuracy), the combinatorial search over architectures, hyperparameters, and implementation strategies is exactly the kind of work that is tedious and error-prone for humans but natural for an autonomous system with enough compute.
The bottleneck to progress in medicine, energy, materials science, and countless other fields is not a shortage of ideas but a shortage of person-hours to test them, and AI systems that could \emph{multiply} human research throughput by orders of magnitude would have enormous impact.

There is a useful analogy from competitive chess: for a decade after Deep Blue, the strongest players were human--AI teams \citep{kasparov2017deep}; research may be in this phase now.
Frontier models can handle literature search, code implementation, hyperparameter sweeps, and failure analysis, but they still benefit from human direction: choosing which problems matter, recognizing misguided lines of inquiry, and providing the judgment that turns throughput into insight.
The question is: \emph{what is the right division of labor, and what infrastructure makes human--AI research collaboration most productive?}

To this end, we introduce \system, an autonomous multi-agent system inspired by Claude Code, designed to extend frontier-model agentic capabilities from software engineering to scientific research through computationally intensive experimental campaigns.
The user provides a dataset and a natural-language objective; the system handles the experimental grind while the human monitors progress and can interact with the Strategist agent to suggest experiments, prune directions, or inject intuitive leaps.
\system operates through three phases: (1)~it \textbf{adapts and explores}, configuring prompts and metrics, writing analysis code, and producing a research report; (2)~it \textbf{builds its own evaluation framework} via an adversarial Builder/Critic/Tester loop; and (3)~it \textbf{experiments at scale} via a Strategist/Worker loop on GPU clusters.
All domain-specific behavior lives in plug-in adapters generated by the model itself, and a persistent \emph{playbook} accumulates domain knowledge across experiments.
We cap each campaign at 50~experiments and run fully hands-off for fair comparison, but the intended workflow is collaborative.

We evaluate \system on three deliberately diverse domains with two frontier LLMs (GPT-5.2 and Claude Opus~4.6).
On \textbf{CUDA kernel optimization} \citep{kernelbench2025}, the system writes custom GPU kernels and benchmarks them against PyTorch's optimized compiler; it produces kernels that run $4.4\times$ faster on average, with the best kernel achieving a $91.4\times$ speedup.
On \textbf{LLM pretraining} \citep{pleias2025synth}, given a 20-minute compute budget the system searches over model architectures and training recipes; the full system achieves 22\% lower validation loss than a single-shot baseline using the same model.
On \textbf{traffic forecasting} \citep{lai2018lstnet}, the system beats seasonal baselines by 23--25\% after researching and implementing published model families from the literature (e.g., PatchTST, TFT, iTransformer).
Baselines for a system like \system are inherently difficult to construct: because the system searches the literature, selects architectures, and tunes hyperparameters as part of a single campaign, there is no single model to compare against.
We therefore compare against other autonomous research systems, simple iterative LLM loops \citep{karpathy2026autoresearch}, single-shot LLM baselines, and published results for the individual model families that \system itself discovers.
Each campaign costs \$150--200 in LLM API calls (not including GPU compute) and completes in 12--48 hours on 4$\times$H100 hardware.

Our contributions:
\begin{enumerate}[nosep]
  \item We introduce \system, an autonomous multi-agent research system with self-generated domain adapters, adversarial evaluation construction, and a persistent playbook for knowledge accumulation, enabling one pipeline to handle qualitatively different domains.
  \item We evaluate across three domains with two frontier LLMs, ablating key components and showing that different models discover complementary solutions.
  \item We open-source the full codebase at \url{https://brendanhogan.github.io/alphalab-paper/} with prompt templates, playbook excerpts, and additional results in the appendix.
\end{enumerate}

\section{System design}
\label{sec:system}

At a high level, \system is a \emph{harness}: a combination of tools and a structured environment that converts a frontier LLM into an autonomous research agent.
We define the harness formally as a tuple $\mathcal{H} = (\mathcal{M}, \mathcal{T}, \mathcal{E})$, where $\mathcal{M}$ is a frontier language model (treated as a black box), $\mathcal{T}$ is a set of tools the model can invoke, and $\mathcal{E}$ is a phased environment that structures the research workflow.
The system is \textbf{LLM-agnostic}: any model that supports tool-use and multi-modal input (agents must be able to read plots and visualizations they generate) can be substituted for $\mathcal{M}$ with no changes to $\mathcal{T}$ or $\mathcal{E}$.
We evaluate with GPT-5.2 and Claude Opus~4.6 in Section~\ref{sec:results}; all differences in outcomes are attributable to the model, not the infrastructure.

\subsection{Tools}
\label{sec:tools}

Every agent in \system has access to the same core tool set $\mathcal{T}$ (full listing in Appendix~\ref{app:tools}).
The three most important tools are: \textbf{shell access} (\texttt{shell\_exec}), which accounts for ${\sim}50\%$ of all tool calls: the agent runs as a user account in a Unix shell and can do anything a human could from a terminal: write code, install packages, run training jobs, manage Slurm submissions, and debug failures; \textbf{web search} (\texttt{web\_search}), used heavily in early phases to research prior work, survey existing approaches, and read library documentation before writing code; and \textbf{sub-agent spawning} (\texttt{spawn\_agent}), which launches a new instance of $\mathcal{M}$ with its own context window and full tool access, enabling recursive delegation without polluting the parent agent's context.

\subsection{Environment: four-phase pipeline}
\label{sec:environment}

The environment $\mathcal{E}$ structures the research workflow into four sequential phases (Figure~\ref{fig:pipeline}); we summarize each below and provide detailed walkthroughs, prompt examples, and case studies in Appendix~\ref{app:system}.
Each phase produces artifacts that are consumed by subsequent phases, and a Supervisor agent monitors health across the pipeline, intervening when error rates spike (Section~\ref{sec:supervisor}).

\subsubsection{Phase~0: Adapter resolution}
\label{sec:adapter}

All domain-specific behavior in \system is parameterized by a \textbf{domain adapter} $\mathcal{A}$: a collection of 11 files comprising a manifest (metric definitions, experiment structure, entry points), 9~prompt templates (one per agent role), and a domain knowledge document that is injected into every agent's context for the entire campaign.

Phase~0 resolves the adapter by resuming from a prior campaign, customizing a built-in template, or, for novel domains, generating all 11~files from scratch by examining the dataset and searching the web for relevant prior work.
The key idea is that \emph{prompt engineering is performed by the model}, grounded in the actual data.
The most important file is \texttt{domain\_knowledge.md}, an artifact prepended to every agent's context for the entire campaign, encoding metric formulas, data quirks, and domain priors so that downstream agents need not re-discover them.

\subsubsection{Phase~1: Data exploration}
\label{sec:explorer}

A single Explorer agent operates autonomously for several hours: it first generates a \texttt{plan.md} checklist, then works through it, writing and running Python scripts, generating plots, searching the web for relevant papers and best practices, and updating its notes after each finding.
It produces two outputs: a human-readable research report (\texttt{data\_report/}) and a machine-readable \texttt{learnings.md} consumed by Phases~2 and~3.
The plan-then-execute structure prevents drift; further detail is in Appendix~\ref{app:explorer}.

\subsubsection{Phase~2: Adversarial evaluation construction}
\label{sec:eval}

Evaluation correctness is the single most important property of an autonomous research system.
If the metric is wrong, every experiment optimizes the wrong objective, the playbook fills with false knowledge, and errors compound silently.
Phase~2 addresses this through a multi-agent adversarial loop:

\begin{equation}
\label{eq:phase2}
\text{Builder} \;\xrightarrow{\text{code}}\; \text{Critic} \;\xrightarrow{\text{review}}\; \begin{cases} \text{Tester} & \text{if no critical issues} \\ \text{Builder} & \text{otherwise} \end{cases}
\end{equation}

The \textbf{Builder} receives Phase~1 learnings and constructs the full evaluation framework (data loading, splitting, metric computation, orchestration).
The \textbf{Critic}, a fresh agent with no shared context, audits for data leakage, lookahead bias, and metric errors.
The \textbf{Tester} writes and runs an automated test suite; the loop terminates when all tests pass or a maximum iteration count $I_{\max}$ is reached.
The cap is necessary because an LLM instructed to find issues will \emph{continue to} find something to flag; in practice, all substantive problems are resolved within 5--10 iterations, after which the remaining findings are stylistic or inconsequential.

\subsubsection{Phase~3: GPU-scale experimentation}
\label{sec:phase3}

Phase~3 is the core of the system: a sustained experimental campaign where the Strategist proposes experiments, Workers execute them on a GPU cluster, and a playbook accumulates knowledge.
A pure-Python Dispatcher orchestrates the process without making any LLM calls itself.
The Dispatcher manages cluster resources through Slurm: it tracks available GPUs, writes and submits Slurm job scripts, monitors job status, and automatically reassigns freed GPUs to the next highest-priority task.
This means the system can keep a multi-node GPU cluster fully utilized around the clock: experiments are queued, dispatched, and monitored without any human involvement, and the cluster never sits idle between experiments.

\paragraph{Experiment lifecycle.}
Each experiment $e$ progresses through a state machine:
\begin{equation}
\label{eq:lifecycle}
e: \quad \texttt{queued} \;\to\; \texttt{implement} \;\to\; \texttt{execute} \;\to\; \texttt{analyze} \;\to\; \texttt{done}
\end{equation}
with a \texttt{fix} state for failed experiments (limited to $k=2$ repair attempts).
The Dispatcher assigns tasks to Workers with a strict priority ordering:
\begin{equation}
\label{eq:priority}
\text{priority}(\texttt{fix}) > \text{priority}(\texttt{analyze}) > \text{priority}(\texttt{implement})
\end{equation}
This ordering maximizes information flow: fixes recover sunk GPU cost, analysis produces debriefs that inform the Strategist, and implementation can always wait.

\paragraph{Strategist.}
Invoked periodically, the Strategist receives the current leaderboard, all recent experiment debriefs, the full playbook, Phase~1 learnings, and remaining budget $B$.
It outputs new experiment specifications, cancellation decisions, and an updated playbook.
Budget management is graduated from broad exploration ($B > 20$) to focused refinement ($B \leq 10$) to stopping ($B = 0$).

\paragraph{Worker.}
Each Worker receives exactly one task: \texttt{implement} (write code, test locally, submit to Slurm), \texttt{analyze} (extract metrics, write a structured debrief), or \texttt{fix} (diagnose failure, patch, resubmit; up to $k{=}2$ attempts).
Workers do not communicate directly; the playbook is the sole channel through which one Worker's findings reach the next.

\paragraph{Playbook.}
The persistent knowledge artifact and central feedback mechanism.
After each Strategist turn, the playbook is updated with compressed findings: what works, what fails, and why.
It is injected into the context of every subsequent Strategist and Worker call, creating a feedback loop that functions as online prompt optimization: by the end of a campaign, it contains domain-specific methodology that did not exist at launch.

\paragraph{Convergence.}
The campaign terminates when no improvement in the primary metric has occurred for $C$ consecutive experiments (default $C = 20$), or the budget $B$ is exhausted (in practice, all campaigns in this paper were budget-limited; see Appendix~\ref{app:convergence}).

\subsubsection{Supervisor}
\label{sec:supervisor}

A meta-agent that validates artifacts between phases and monitors Phase~3 health.
The Supervisor is triggered when the error rate exceeds a threshold $\tau = 0.4$ over a sliding window:
\begin{equation}
\label{eq:supervisor}
\frac{|\{e \in W : e.\text{status} = \texttt{failed}\}|}{|W|} > \tau \implies \text{trigger Supervisor}
\end{equation}
When triggered, the Supervisor reads recent failure logs, diagnoses systemic issues, and patches the adapter's \texttt{domain\_knowledge.md}, committing the change to git so it is traceable and reversible.
Examples of Supervisor interventions are detailed in the appendix.

\section{Experiments and Results}
\label{sec:results}

We refer to a full end-to-end pipeline execution as a \emph{campaign}, and each individual architecture or configuration tested within a campaign as an \emph{experiment}.
Every meaningful campaign takes approximately 40 hours and costs \$150--200 in LLM API calls (GPU compute costs are additional and depend on infrastructure), making exhaustive ablation prohibitive; all single-run comparisons are indicative rather than conclusive.
To ensure apples-to-apples comparison, we run Phase~2 once and copy the harness to all runs within a domain.
Every domain runs with GPT-5.2 and Claude Opus~4.6, all other variables held constant.
As discussed in Section~\ref{sec:intro}, we compare against a greedy loop baseline \citep{karpathy2026autoresearch}, a single-shot LLM baseline, and published results for model families that \system itself discovers.
Hardware: 4$\times$H100 NVL 80\,GB; budget: 50~experiments per campaign.
The greedy loop baseline is given the same 50-iteration budget; the single-shot baseline is intentionally minimal (one call with one retry) and serves as a lower bound.

\subsection{LLM pretraining (speedrun)}
\label{sec:llm-results}

\paragraph{Task.}
Inspired by Karpathy's NanoGPT speedrun \citep{jordan2024moddednanogpt}, the agent must minimize validation bits-per-byte (\texttt{val\_bpb}) on the PleIAs SYNTH corpus \citep{pleias2025synth} when training a $<$100M parameter language model from scratch under a 20-minute wall-clock budget.
The optimization space spans architecture (depth, width, attention/FFN variants, normalization, positional encoding), optimizer (AdamW with various schedules), and training dynamics.

\paragraph{Main results.}
Table~\ref{tab:llm-results} summarizes results across three models and two baselines.
Opus~4.6 achieves a best val\_bpb of \textbf{0.7578}, substantially outperforming GPT-5.2 (0.9697) and Sonnet~4.6 (0.8686), a 22\% improvement over GPT-5.2.
Sonnet~4.6 falls between the two at 0.8686, converging on a distinct recipe (11L$\times$768d with a Muon/AdamW hybrid optimizer and QK-norm) that neither other model discovered.
GPT-5.1-mini, a smaller and cheaper model, failed entirely: it ran 33~experiments but could not correctly implement the nats-to-BPB conversion, producing values ranging from 0 to $\infty$; it self-diagnosed the issue in its own playbook but only after exhausting the budget.
The gap between Opus and GPT-5.2 is driven primarily by GPT-5.2's 38\% experiment failure rate (PyTorch API breaking changes) and Opus's convergence on wider-shallower architectures (10--12 layers, 672--752-dim) better suited to the $<$100M regime.

\begin{table}[t]
\centering
\small
\begin{tabular}{@{}lccc@{}}
\toprule
\textbf{Model} & \textbf{Best val\_bpb} & \textbf{Best config} & \textbf{Cost} \\
\midrule
\system + GPT-5.2    & 0.9697 & 8L$\times$512d, GQA, cosine      & $\sim$\$150 \\
\system + Sonnet~4.6 & 0.8686 & 11L$\times$768d, QK-norm, Muon   & $\sim$\$120 \\
\system + Opus~4.6   & \textbf{0.7578} & 10L$\times$752d, QK-norm & $\sim$\$200 \\
\system + GPT-5.1-mini & ---$^\dagger$ & (no valid results) & $\sim$\$40 \\
\midrule
Greedy loop (GPT-5.2) & 1.020 & 12L$\times$768d, LLaMA, AdamW & $\sim$\$50 \\
Single-shot (GPT-5.2) & 1.248 & 27.4M LLaMA-style & $<$\$1 \\
\bottomrule
\end{tabular}
\caption{\textbf{LLM pretraining speedrun results.}  Task: train a $<$100M-parameter language model from scratch on the PleIAs SYNTH corpus under a 20-minute wall-clock budget on a single H100, then measure validation bits-per-byte (val\_bpb, lower is better) on a held-out set.  Configs describe the best architecture found: ``L'' = layers, ``d'' = model dimension, ``GQA'' = grouped-query attention, ``QK-norm'' = query/key normalization for training stability, ``cosine''/``Muon'' = learning rate schedule or optimizer.  The greedy loop is a Karpathy-style AutoResearch baseline \citep{karpathy2026autoresearch}: the LLM proposes a modification, trains, and keeps or reverts based on the metric (50 iterations, 3 improvements).  The single-shot baseline is one LLM call with one retry.  $^\dagger$GPT-5.1-mini ran 33 experiments but failed to correctly implement the BPB metric; all results were broken (see text).}
\label{tab:llm-results}
\end{table}

\paragraph{Ablations.}
Because this domain has the fastest iteration cycle, we concentrate all ablations here.
Each ablation modifies a single pipeline component while holding everything else constant (same hardware, 50-experiment budget, shared Phase~2 harness).
Table~\ref{tab:ablation} summarizes results.

\begin{table}[t]
\centering
\small
\begin{tabular}{@{}llcc@{}}
\toprule
\textbf{Variant} & \textbf{Removed} & \textbf{Best val\_bpb} & \textbf{$\Delta$ vs.\ full} \\
\midrule
Full \system       & (nothing)          & 0.9697 & ---                  \\
Skip Phase~1       & Exploration        & 1.0908 & $+$0.121 ($+$12.5\%) \\
No playbook        & Knowledge acc.     & 0.9941 & $+$0.024 ($+$2.5\%)  \\
\midrule
Variance (5 runs)  & (nothing)          & $0.994 \pm 0.025$ & ---          \\
\bottomrule
\end{tabular}
\caption{\textbf{Ablation results on LLM speedrun} (GPT-5.2, 50-experiment budget).  Each ablation removes a single pipeline component while holding everything else constant (same hardware, shared Phase~2 harness).  ``Skip Phase~1'' removes the data exploration phase where the system researches prior work on arXiv and analyzes the dataset before experimenting.  ``No playbook'' removes the persistent knowledge document that accumulates lessons across experiments; without it, the system cannot remember what worked or failed in earlier experiments and must rediscover these lessons from scratch.  The ``Variance'' row reports the mean $\pm$ standard deviation of the best val\_bpb across five runs of the full system.}
\label{tab:ablation}
\end{table}

\textbf{Skip Phase~1 (no exploration):} Bypassing data exploration costs 0.121 BPB ($+$12.5\%), the largest single-component degradation.
Without prior analysis, the Strategist wastes early experiments on configurations that Phase~1 would have flagged as suboptimal (e.g., not discovering that including the query prefix improves BPB, or that byte-level tokenization enables direct BPB computation).

\textbf{No playbook:} With the playbook disabled, the system achieves 0.9941, 2.5\% worse than the full system.
Workers without playbook context must rediscover lessons (e.g., ``disable torch.compile'', ``use byte tokenization'') that earlier experiments had already established, wasting budget on known-bad configurations.

\paragraph{Phase~3 variance.}
\label{sec:variance}
To assess how much inter-model difference reflects genuine model quality versus search stochasticity, we ran GPT-5.2's Phase~3 five times with identical inputs (same harness, learnings, data), varying only the stochastic LLM generation, the primary run plus four additional replications.
The five runs yield val\_bpb of 0.964, 0.970, 1.006, 1.011, and 1.020, a spread of 0.056 BPB from identical inputs, comparable in magnitude to the gap between the greedy baseline (1.020) and the full system's best (0.964).
All five runs match or beat the greedy loop baseline (1.020), suggesting that \system's advantage over na\"ive iteration is robust to search stochasticity.

\subsection{CUDA kernel optimization}
\label{sec:cuda-results}

\paragraph{Task.}
We evaluate on KernelBench \citep{kernelbench2025}, extended by Sakana AI's CUDA Engineer benchmark \citep{sakanakernelbench2025}.
The task: given a PyTorch operator, write an optimized CUDA kernel that is both \emph{correct} (matches the reference output) and \emph{fast} (exceeds \texttt{torch.compile} performance).
We target Level~1 (100 single-operator problems) and Level~2 (100 fusion patterns).
We define $\text{fast}_p$ as the fraction of correct kernels achieving $>p\times$ speedup over \texttt{torch.compile}; $\text{fast}_1$ denotes correct \emph{and} faster than compiled PyTorch.

\paragraph{Results.}
Table~\ref{tab:cuda-results} presents results at three levels of comparison.
On the 54 tasks where both models produced correct kernels (same tasks, hardware, baseline), GPT-5.2 leads with 5.17$\times$ mean speedup vs Opus's 4.63$\times$.
The table also places \system in the context of prior work, though external comparisons are approximate due to differences in hardware and baseline.
\system did not attempt all 200 KernelBench tasks; each campaign was budget-capped at 50 experiments for consistency across domains.
Per-level breakdowns and per-kernel comparisons with Sakana AI are in the appendix.

\begin{table}[t]
\centering
\small
\begin{tabular}{@{}llccc@{}}
\toprule
\textbf{System} & \textbf{Model} & \textbf{Correct} & \textbf{Mean spd.} & $\textbf{fast}_1$ \\
\midrule
\multicolumn{5}{@{}l}{\textit{Direct comparison (54 tasks, both models correct, H100, vs \texttt{torch.compile})}} \\
\system & GPT-5.2  & 54 & \textbf{5.17$\times$} & \textbf{91\%} \\
\system & Opus~4.6 & 53 & 4.63$\times$ & 70\% \\
\midrule
\multicolumn{5}{@{}l}{\textit{Full run (H100, vs \texttt{torch.compile})}} \\
\system & GPT-5.2  & 110/119 & 4.40$\times$ & 83\% \\
\system & Opus~4.6 &  76/87  & 4.00$\times$ & 70\% \\
\midrule
\multicolumn{5}{@{}l}{\textit{External baselines (different hardware and/or baseline -- approximate comparison)}} \\
\system (H100, vs \texttt{native})  & GPT-5.2  & 110/119 & 3.47$\times$ & 75\% \\
\system (H100, vs \texttt{native})  & Opus~4.6 &  76/87  & 3.27$\times$ & 87\% \\
Sakana AI (H100, vs \texttt{native}) & Ensemble & 190/200 & 1.49$\times$ & --- \\
KernelBench (L40S, vs \texttt{native}) & R1 (10 calls) & --- & --- & 43/72\% \\
KernelBench (L40S, vs \texttt{native}) & R1 (1-shot) & --- & --- & 12/36\% \\
\bottomrule
\end{tabular}
\caption{\textbf{CUDA kernel optimization results.}  Task: write a CUDA kernel matching PyTorch's output but faster, on KernelBench \citep{kernelbench2025} (100 single-op + 100 fusion tasks).  \textit{Top:} head-to-head on 54 tasks where both models produced correct kernels -- same tasks, hardware, baseline.  \textit{Middle:} each model's full campaign (budget-capped at 50 experiments, so $<$200 tasks attempted).  \textit{Bottom:} \system re-reported against \texttt{torch.native} to enable comparison with external baselines.  Sakana AI \citep{sakanarobust2025}: 5-model ensemble, excludes contaminated tasks.  KernelBench: L40S GPU, $\text{fast}_1$ as L1/L2\%.  Cross-system comparison is approximate due to hardware and baseline differences.}
\label{tab:cuda-results}
\end{table}

\subsection{Traffic forecasting}
\label{sec:traffic-results}

\paragraph{Task.}
Hourly road occupancy forecasting for 862 San Francisco Bay Area freeway sensors \citep{lai2018lstnet}, predicting 24~hours ahead.
Occupancy values lie in $[0,1]$ with strong daily and weekly seasonality.
Primary metric: RMSE (minimize).
Baseline: Seasonal Na\"ive(168) (repeat the value from one week ago; see Table~\ref{tab:traffic-results}), RMSE $\approx$ 0.0287.

\paragraph{Results.}
Table~\ref{tab:traffic-results} summarizes results.
Opus~4.6 achieves the best RMSE of \textbf{0.02142} ($-$25\%) via TFT \citep{tft} (an attention-based multi-horizon forecasting architecture with recurrence ); GPT-5.2 reaches 0.02204 ($-$23\%) via iTransformer \citep{liu2024itransformer} (a Transformer variant treating each sensor as a token).
The two models searched differently: Opus converged entirely on TFT variants, while GPT-5.2 explored iTransformer, PatchTST \citep{Yuqietal-2023-PatchTST}, TSMixer \citep{chen2023tsmixer}, and N-HiTS \citep{NHITS}.
The table also reports literature RMSE values alongside \system's own reimplementations; \system's tuned versions consistently outperform literature defaults, suggesting the gap reflects hyperparameter tuning rather than architectural novelty.
Both LLM baselines beat the seasonal na\"ive but remain substantially worse than \system.
Notably, the greedy loop performs slightly worse than the single-shot baseline on this domain, likely due to path dependence: the greedy loop's sequential propose-and-revert structure is constrained by its starting point and can converge to a local optimum, whereas the single-shot call happened to produce a stronger initial architecture.
Detailed top-$K$ tables are in the appendix.

\begin{table}[t]
\centering
\small
\begin{tabular}{@{}lccc@{}}
\toprule
\textbf{Model} & \textbf{Lit.\ RMSE} & \textbf{\system RMSE} & \textbf{Cost} \\
\midrule
\system + Opus~4.6 (TFT, dropout 0.3) & --- & \textbf{0.02142} & $\sim$\$200 \\
\system + GPT-5.2 (iTransformer ctx336) & --- & 0.02204 & $\sim$\$180 \\
\midrule
DeepAR \citep{DBLP:journals/corr/FlunkertSG17} &  0.0249 & 0.0251 & --- \\
PatchTST \citep{Yuqietal-2023-PatchTST} & 0.0311 & 0.0226 & --- \\
TiDE \citep{das2023longterm} & 0.0281 & 0.0230 & --- \\
DLinear \citep{Zeng_Chen_Zhang_Xu_2023} & 0.0335 & 0.0234 & --- \\
SeasonalNaive (168) & 0.0287 & 0.0287 & --- \\
Seasonal Average & 0.0341 & --- & --- \\
Weighted Ensemble & 0.0346 & --- & --- \\
Theta \citep{ASSIMAKOPOULOS2000521}       & 0.0370 & ---  & --- \\
ETS \citep{hyndman2021forecasting}    & 0.0404 & ---  & --- \\
\midrule
Single-shot (GPT-5.2) & --- & 0.02686 & $<$\$1 \\
Greedy loop (GPT-5.2) & --- & 0.02779 & $\sim$\$50 \\
\bottomrule
\end{tabular}
\caption{\textbf{Traffic forecasting results} (RMSE, lower is better).  Task: predict hourly road occupancy 24h ahead for 862 freeway sensors.  Seasonal Na\"ive(168) repeats the value from one week ago (0.0287).  ``Lit.\ RMSE'' = published result using each paper's protocol; ``\system RMSE'' = best result when \system independently discovered and tuned that model family on its own harness.  Discrepancies reflect tuning and protocol differences.  Single-shot = one LLM call; greedy loop = 50 sequential propose-then-train iterations.  The greedy loop's worse-than-single-shot result reflects path dependence: its sequential propose-and-revert structure can converge to a local optimum that a single lucky call avoids.}
\label{tab:traffic-results}
\end{table}

\subsection{What the system discovered}
\label{sec:discovered}

The playbook is perhaps the most interesting artifact the system produces.
It starts empty; by the end of a campaign it contains domain-specific methodology that did not exist anywhere in \system's prompts or code at launch, written by the Strategist, experiment by experiment, from results.
We highlight representative excerpts (verbatim, unedited):

\textbf{CUDA kernels} (GPT-5.2): \emph{``Diagonal matrix ops: $\text{diag}(A) \cdot B$ is just elementwise multiply of the diagonal vector with each row of $B$.  Yields 10--68$\times$ over PyTorch's full matmul.  Warp-shuffle reductions yield 73--75$\times$ on sum operations.  Do not attempt convolution kernels: cuDNN is too well-optimized; handwritten runs at 0.05--0.73$\times$.''}

\textbf{LLM pretraining} (Opus~4.6): \emph{``Wider-shallower architectures (10--12 layers, 672--752-dim) outperform deeper-narrower at this parameter budget.  QK-norm stabilizes training at high learning rates.''}
GPT-5.2's playbook emphasized different findings: \emph{``Disable torch.compile: compilation overhead is significant under a 20-min budget.  GQA reduces memory and enables larger batch sizes.''}

\textbf{Traffic} (GPT-5.2): \emph{``Per-horizon affine calibration on top of iTransformer predictions is the single most impactful post-processing step: adding it consistently pushes RMSE below 0.023 where the raw model sits above it.''}

The ``do not attempt'' entries are as valuable as positive findings, preventing budget waste on disproven approaches.
Additional playbook excerpts are in the appendix.

\subsection{Discussion}
\label{sec:exp-discussion}

GPT-5.2 leads on CUDA kernels while Opus leads on LLM pretraining and narrowly on traffic; neither dominates uniformly (Table~\ref{tab:summary}, Appendix).
The models discover qualitatively different solutions in every domain, reinforcing the case for multi-model campaigns.
Even within the same model family, Opus and Sonnet discover qualitatively different architectures on LLM pretraining (Section~\ref{sec:llm-results}).
On the other end of the capability spectrum, GPT-5.1-mini failed to produce any trustworthy results on LLM pretraining: it could not maintain the engineering rigor needed to correctly implement metrics and debug environment failures (Appendix~\ref{app:repro}), suggesting a minimum model capability threshold for autonomous research.

\section{Related work}
\label{sec:related}

The AI Scientist~\citep{lu2024aiscientist,lu2025aiscientistv2}, AIDE~\citep{jiang2025aide}, and Agent Laboratory~\citep{schmidgall2025agentlab} demonstrated autonomous research with varying degrees of scope; multi-agent frameworks such as MetaGPT~\citep{hong2023metagpt}, ChatDev~\citep{qian2024chatdev}, and AutoGen~\citep{wu2023autogen} introduced role-based decomposition for software tasks.
\system's roles target research campaigns rather than software projects, and its playbook provides cross-experiment memory absent from these systems.
Voyager~\citep{wang2023voyager} is the closest analog to the playbook (a skill library accumulated during play); Reflexion~\citep{shinn2023reflexion} maintains verbal memory within a single task.
FunSearch~\citep{romera2024funsearch}, AlphaCode~\citep{li2022alphacode}, GPU Kernel Scientist~\citep{gpukernelscientist2025}, and Sakana AI's CUDA Engineer~\citep{sakanakernelbench2025} apply LLM-guided search to specific domains.
AutoML systems~\citep{feurer2015autosklearn,erickson2020autogluon,akiba2019optuna} search predefined configuration spaces; \system constructs its own evaluation methodology and reasons about why experiments succeed or fail.
The broader trend toward harness engineering \citep{poetiq2025arcagi} motivates \system's design.

\section{Conclusion}
\label{sec:conclusion}

We presented \system, an autonomous multi-agent system that automates the experimental research cycle across qualitatively different domains at a cost of \$150--200 per campaign.
The central finding is that different frontier models discover different solutions in every domain, suggesting multi-model campaigns provide complementary search coverage.
Key limitations include premature playbook convergence (explicit diversity budgets are needed), environmental fragility (PyTorch API changes caused 38\% failure rates), single-run comparisons, and the lack of sandboxing for LLM-generated code.
We believe research is entering a phase where human--AI teams outperform either alone, and we hope open-sourcing \system accelerates progress toward that vision.

\section*{Acknowledgements}

We are especially thankful to CoreWeave, whose purpose-built AI cloud platform powered our experiments.

\newpage
\bibliography{colm2026_conference}
\bibliographystyle{colm2026_conference}

\newpage

\begin{center}
{\Large\textbf{LLM Disclosure Statement}}
\end{center}

\vspace{1em}

\noindent In accordance with COLM policy, we disclose the following uses of LLMs in the preparation of this work.

\paragraph{System under study.}
\system is an LLM-powered autonomous research system; all experimental code, evaluation frameworks, domain adapters, and playbook content reported in this paper were generated by the frontier LLMs under evaluation (GPT-5.2, Claude Opus~4.6, Claude Sonnet~4.6) as part of the system's normal operation.

\paragraph{System development.}
The \system codebase was built with the assistance of agentic coding tools (Claude Code).
The core idea and system architecture were human-conceived; detailed design decisions, implementation, and debugging were carried out iteratively with LLM assistance.

\paragraph{Figures.}
The main pipeline figure (Figure~\ref{fig:pipeline}) was generated using Nano Banana Pro (Gemini~3 Pro Image) from Google.

\paragraph{Analysis and writing.}
Experimental results were analyzed and interpreted by the authors.
The manuscript was outlined and rough-drafted by the authors, then iteratively refined with LLM assistance for clarity and presentation.
All scientific claims and conclusions are the sole responsibility of the authors.

\appendix
\newpage

\begin{center}
{\Large\textbf{Appendix}}
\end{center}

\vspace{1em}

\begin{tabular}{@{}lp{0.72\linewidth}@{}}
\textbf{Appendix~\ref{app:system}} & \textbf{System details} -- Full tool set, domain adapter file structure, detailed walkthroughs of all four phases with prompt examples, and Supervisor intervention case studies. \\[4pt]
\textbf{Appendix~\ref{app:results}} & \textbf{Extended experimental results} -- Cross-domain summary, per-model top-$K$ and full experiment tables for LLM speedrun and traffic, per-level CUDA breakdown, per-kernel Sakana AI comparison, and convergence curves. \\[4pt]
\textbf{Appendix~\ref{app:playbook}} & \textbf{Playbook excerpts and dynamics} -- Verbatim playbook excerpts from all domains, growth trajectory analysis, self-correction events, and cross-model comparison of playbook styles. \\[4pt]
\textbf{Appendix~\ref{app:ui}} & \textbf{User interface} -- Description of the real-time web dashboard (Kanban board, leaderboard, file viewer, conversation stream, human-in-the-loop chat). \\[4pt]
\textbf{Appendix~\ref{app:additional}} & \textbf{Additional experiments} -- Financial time series forecasting (exchange rates) with auto-generated adapter. \\[4pt]
\textbf{Appendix~\ref{app:cost}} & \textbf{Cost and token breakdown} -- Token usage per campaign, phase-level distribution, cost per experiment, and model cost comparison. \\[4pt]
\textbf{Appendix~\ref{app:failures}} & \textbf{Failure analysis} -- Three-tier failure taxonomy (programmatic, evaluation, strategic) with per-domain breakdown and error rates. \\[4pt]
\textbf{Appendix~\ref{app:repro}} & \textbf{Reproducibility details} -- Model versions, API parameters, hardware specs, repository structure, and variance experiment protocol. \\
\end{tabular}
\newpage
\section{System details}
\label{app:system}

\subsection{Full tool set}
\label{app:tools}

\begin{table}[h]
\centering
\small
\caption{Full tool set available to all \system agents.}
\begin{tabular}{@{}lp{0.68\linewidth}@{}}
\toprule
\textbf{Tool} & \textbf{Description} \\
\midrule
\texttt{shell\_exec} & Full shell access: execute arbitrary commands, write and run code, install packages, manage files. \\
\texttt{read\_file} & Read any file in the workspace. \\
\texttt{grep\_file} & Search file contents by pattern. \\
\texttt{web\_search} & Search the internet for papers, documentation, and best practices. \\
\texttt{view\_image} & View plots and visualizations the agent has generated. \\
\texttt{spawn\_agent} & Launch a sub-agent: a new instance of $\mathcal{M}$ with its own context and full tool access. \\
\texttt{read\_board} & Read the current experiment leaderboard (Phase~3). \\
\texttt{update\_playbook} & Append entries to the persistent playbook (Phase~3). \\
\texttt{propose\_experiment} & Submit a new experiment specification (Phase~3, Strategist only). \\
\texttt{report\_to\_user} & Send a status report to the human operator. \\
\bottomrule
\end{tabular}
\end{table}

Across a representative LLM speedrun campaign (GPT-5.2, 50~experiments, 2{,}892 API calls), tool usage breaks down as follows: \texttt{shell\_exec} accounts for ${\sim}49.5\%$ of all tool calls, reflecting the code-heavy nature of the work; \texttt{read\_file} for ${\sim}21.8\%$ (reading experiment outputs, debriefs, and framework code); \texttt{grep\_file} for ${\sim}12.3\%$ (searching for errors, patterns, and configurations); \texttt{web\_search} for ${\sim}8.1\%$ (concentrated in Phase~1); and the remaining ${\sim}8.3\%$ split across \texttt{propose\_experiment}, \texttt{update\_playbook}, \texttt{read\_board}, \texttt{view\_image}, and \texttt{report\_to\_user}.

Tool usage varies substantially by phase.
Phase~1 (exploration) is dominated by \texttt{shell\_exec} (${\sim}60\%$) and \texttt{web\_search} (${\sim}20\%$), as the agent writes analysis scripts and researches the domain.
Phase~2 (evaluation construction) shifts toward \texttt{read\_file} (${\sim}35\%$) and \texttt{shell\_exec} (${\sim}45\%$), with the Critic reading code and the Tester running test suites.
Phase~3 (experimentation) has the most diverse tool usage: the Strategist uses \texttt{read\_board}, \texttt{propose\_experiment}, and \texttt{update\_playbook} almost exclusively, while Workers are heavily \texttt{shell\_exec}-dominated (${\sim}70\%$).

\subsection{Domain adapter file structure}
\label{app:adapter-detail}

The 11 adapter files are:

\begin{enumerate}[nosep]
  \item \texttt{manifest.yaml} -- metric definitions, direction (min/max), experiment structure, entry points
  \item \texttt{domain\_knowledge.md} -- injected into every agent's context; written by Phase~0 after examining data
  \item \texttt{phase1\_explorer.md} -- prompt for the Explorer agent
  \item \texttt{phase2\_builder.md} -- prompt for the Builder
  \item \texttt{phase2\_critic.md} -- prompt for the Critic
  \item \texttt{phase2\_tester.md} -- prompt for the Tester
  \item \texttt{phase3\_strategist.md} -- prompt for the Strategist
  \item \texttt{phase3\_worker.md} -- prompt for the Worker (implement/analyze/fix)
  \item \texttt{phase3\_supervisor.md} -- prompt for the Supervisor
  \item \texttt{phase0\_customizer.md} -- prompt for the adapter customization agent
  \item \texttt{phase0\_generator.md} -- prompt for full adapter generation from scratch
\end{enumerate}

\noindent Complete adapter files for all domains are available in the code repository.

While all adapters share the same 11-file structure, their contents differ substantially to reflect each domain's evaluation methodology, optimization landscape, and failure modes.

\paragraph{Metric and framework differences.}
The time series adapter defines RMSE as its primary metric (minimize), uses a walk-forward backtesting harness with embargo periods, and requires each experiment to implement a \texttt{Strategy} subclass with \texttt{fit()}/\texttt{predict()} methods.
The CUDA kernel adapter defines speedup over \texttt{torch.compile} as its primary metric (maximize), uses a benchmark harness that checks correctness via \texttt{torch.allclose} and measures median runtime over 100 runs, and requires each experiment to produce a \texttt{kernel.cu} file.
The LLM speedrun adapter defines \texttt{val\_bpb} (minimize), uses a training harness that enforces a 20-minute wall-clock budget and $<$100M parameter constraint, and requires a modified \texttt{train.py}.

\paragraph{Domain knowledge examples.}
The \texttt{domain\_knowledge.md} file is the highest-value adapter artifact because it is injected into every agent's context for the entire campaign.
After Phase~0 customization, these files contain task-specific guidance grounded in the actual data.
Two representative excerpts:

\medskip
\noindent\textbf{LLM speedrun} \texttt{domain\_knowledge.md} (excerpt, after Phase~0 customization):

\begin{quote}\small\itshape
``Primary Metric: \texttt{val\_bpb} (Validation Bits-Per-Byte). Definition: $\text{bpb} = \text{loss\_nats} \times \log_2(e) / \text{bytes\_per\_token}$. [\ldots]
Dataset: PleIAs SYNTH, 500 parquet shards (${\sim}$220\,GB). Languages: en ${\sim}$80.7\%; exercises: memorization ${\sim}$90.6\%. [\ldots]
Architecture sweet spots under 100M: 6--12 layers, d\_model 512--768, RMSNorm + RoPE + SwiGLU.
Sampling is mandatory: reading the full 220\,GB corpus is not feasible within 20 minutes.''
\end{quote}

\noindent\textbf{Traffic forecasting} \texttt{domain\_knowledge.md} (excerpt):

\begin{quote}\small\itshape
``862 freeway sensors, hourly occupancy $[0, 0.724]$, prediction\_length=24.
Test set: 6034 entries = 862 sensors $\times$ 7 rolling windows.
Strong daily seasonality (peak/trough ratio ${\approx}$9.6$\times$); weekly seasonality (weekday mean ${\approx}$1.42$\times$ weekend).
Sensors are correlated: mean pairwise correlation ${\approx}$0.56.
Context length $\geq$192h to access weekly lag.''
\end{quote}

\noindent The same file structure thus encodes fundamentally different domain knowledge (from tokenizer semantics and parameter budgeting (LLM speedrun) to seasonal decomposition and cross-sensor correlation structure (traffic)), while downstream agents consume them through an identical interface.

\subsection{Phase~0: Adapter resolution -- detailed workflow}
\label{app:phase0}

Phase~0 resolves the domain adapter through one of three paths, selected automatically based on the configuration:

\paragraph{Path 1: Resume.}
If \texttt{\{workspace\}/adapter/manifest.json} exists, the adapter is loaded directly.
This enables resuming campaigns without re-running adapter resolution.

\paragraph{Path 2: Customize built-in.}
When the \texttt{domain} field matches a built-in adapter name (\texttt{time\_series}, \texttt{cuda\_kernel}, \texttt{llm\_speedrun}), the template is copied to the workspace and a customization agent examines the actual dataset.
The agent has access to \texttt{shell\_exec}, \texttt{read\_file}, \texttt{read\_adapter}, and \texttt{patch\_adapter\_file}.
It reads a sample of the data, runs exploratory queries (e.g., checking column names, value distributions, data size), and patches the adapter files (especially \texttt{domain\_knowledge.md}) to reflect the specific task.

For example, on the LLM speedrun task, the customization agent sampled 6 parquet shards, discovered the byte-level tokenization scheme (vocab=256), measured the language distribution (${\sim}$80.7\% English), and computed text length statistics.
It then wrote these findings into \texttt{domain\_knowledge.md}, including the exact BPB formula, sampling guidance (``reading the full 220\,GB corpus is not feasible within 20 minutes''), and architecture recommendations grounded in the parameter constraint.

\paragraph{Path 3: Generate from scratch.}
When the \texttt{domain} field contains free-text (e.g., ``forecast exchange rates''), a generation agent creates all 11 adapter files from scratch.
The agent has access to \texttt{write\_adapter\_file} and \texttt{read\_reference\_adapter} (to examine built-in adapters as format references).
It examines the dataset, searches the web for relevant papers and benchmarks, and produces the complete adapter: manifest, domain knowledge, and all 9 prompt templates.
This path is used for novel domains not covered by built-in adapters.

\subsection{Phase~1: Data exploration -- detailed workflow}
\label{app:explorer}

We illustrate Phase~1 with the LLM speedrun campaign (GPT-5.2).
The Explorer begins by creating a structured \texttt{plan.md} checklist, then works through it autonomously:

\begin{quote}\small
\begin{verbatim}
# Plan — PleIAs SYNTH LLM pretraining quality speedrun
- [x] 0. Workspace + environment reconnaissance (GPU/CPU, libs)
- [x] 1. Dataset recon: schema, language distribution, lengths
- [x] 2. Build fast dataloader/tokenizer pipeline
- [x] 3. Define evaluation metric: val_bpb computation
- [x] 4. Implement baseline model/training loop
- [x] 5. Parameter-budgeted model family search (<100M)
- [x] 6. Optimization search: AdamW vs alternatives, schedules
- [x] 7. Architecture tweaks: GPT-2 vs LLaMA-style
- [x] 8. Data curriculum: include query? filter languages?
- [x] 9. Throughput tuning: batch size, compilation
- [x] 10. Run controlled ablations; keep scoreboard
- [x] 11. Assemble deliverables
\end{verbatim}
\end{quote}

\noindent The Explorer wrote and executed 7 Python scripts in \texttt{scripts/} (dataset reconnaissance, tokenizer benchmarking, baseline training, architecture search), generated distribution plots in \texttt{plots/} (language distribution, text length histograms), and produced three deliverables in \texttt{data\_report/}: \texttt{schema.md}, \texttt{statistics.md}, and \texttt{findings.md}.

Key discoveries recorded in \texttt{learnings.md} (verbatim excerpts):
\begin{quote}\small\itshape
``Byte-level tokenization (UTF-8 bytes) enables direct bpb computation (1 token == 1 byte).
Including query as a prefix (Q: \ldots A: \ldots) materially improved short-run val\_bpb (${\sim}$2.46 vs ${\sim}$3.00 at ${\sim}$90s).
seq\_len=1024 performed similarly to 512 in short runs; prefer 1024 for longer-budget training.''
\end{quote}

\paragraph{Traffic forecasting (GPT-5.2).}
The traffic Explorer produced a 20-item plan organized into 7~sections (setup, schema, temporal structure, statistical profiling, seasonality/autocorrelation, cross-sensor dependency, and feature relationships):

\begin{quote}\small
\begin{verbatim}
# Plan — Traffic (GluonTS) Dataset Deep Dive
## Data loading & schema exploration
- [x] Parse train/test JSON, confirm 862 sensors, ~14,036h
- [x] Validate hourly frequency, prediction_length=24
## Target dynamics: seasonality, autocorrelation, stationarity
- [x] ACF/PACF globally and per-cluster (sample sensors)
- [x] Spectral analysis / periodogram (24h, 168h expected)
- [x] Decomposition (STL) on representative sensors
## Cross-sensor dependency & covariance structure
- [x] PCA on standardized series
- [x] Clustering sensors by daily pattern embeddings
\end{verbatim}
\end{quote}

\noindent Key discoveries: 862~sensors with strong dual seasonality (lag-24 ACF=0.87, lag-168 ACF=0.95; weekly periodicity stronger than daily); cross-sensor mean correlation 0.57; PCA PC1 explains 55\% of variance; baseline RMSE of seasonal naive(168) $\approx$~0.0287; context lengths $\geq$192h needed to capture weekly patterns.

\paragraph{CUDA kernels (GPT-5.2).}
The CUDA Explorer's plan focused on benchmarking infrastructure rather than data analysis: inventorying all 200~tasks (100 Level-1 single-op, 100 Level-2 fused-op), profiling tensor sizes ($10^3$ to $>10^8$ elements), and establishing the optimization priority hierarchy.
Key learnings recorded:

\begin{quote}\small\itshape
``200 tasks total: 100 Level~1 (single-op) + 100 Level~2 (fused ops).
Tensor sizes range $10^3$ to $>10^8$ elements.
Level~1: mixed ops (conv, elementwise, reduction, matmul).
Level~2: conv-heavy and matmul-heavy fused patterns.
CUDA compilation takes minutes; subprocess evaluation adds overhead.
Memory hierarchy optimization (coalescing $\to$ tiling $\to$ vectorization) should be the first priority for every kernel.''
\end{quote}

\begin{figure}[h]
\centering
\begin{subfigure}[b]{0.32\textwidth}
\includegraphics[width=\textwidth]{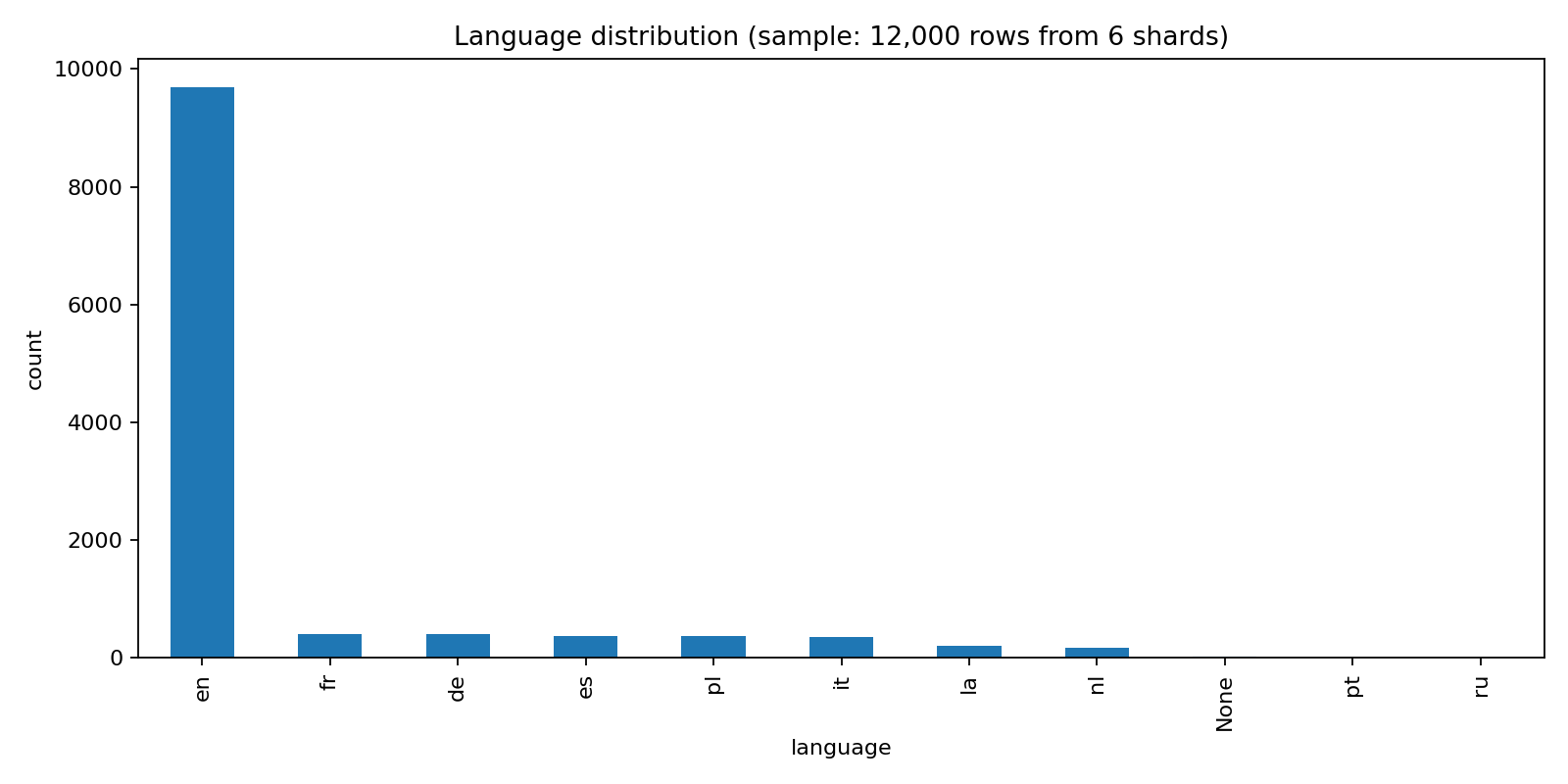}
\caption{LLM: language distribution.}
\end{subfigure}
\hfill
\begin{subfigure}[b]{0.32\textwidth}
\includegraphics[width=\textwidth]{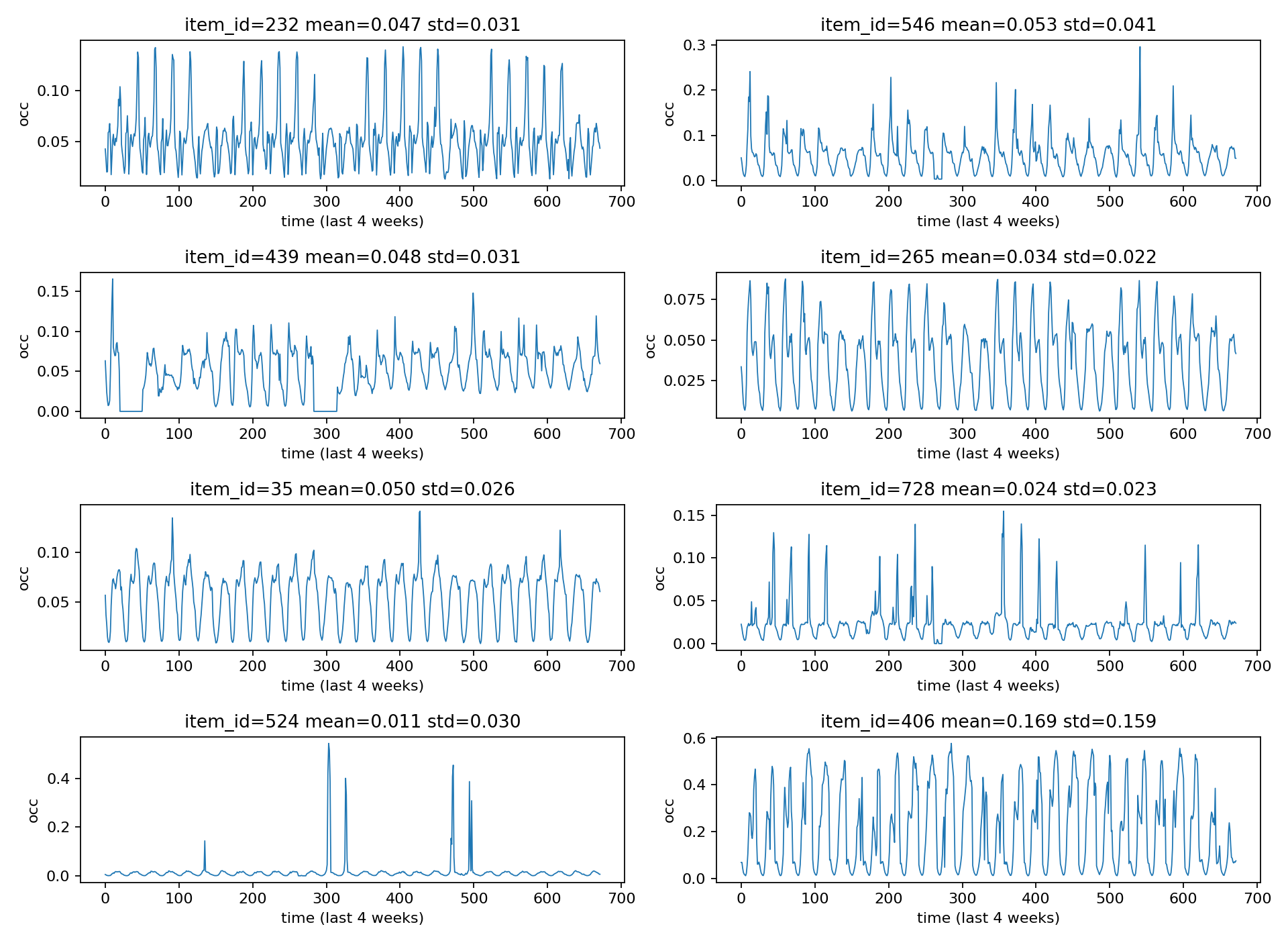}
\caption{Traffic: sample sensor series.}
\end{subfigure}
\hfill
\begin{subfigure}[b]{0.32\textwidth}
\includegraphics[width=\textwidth]{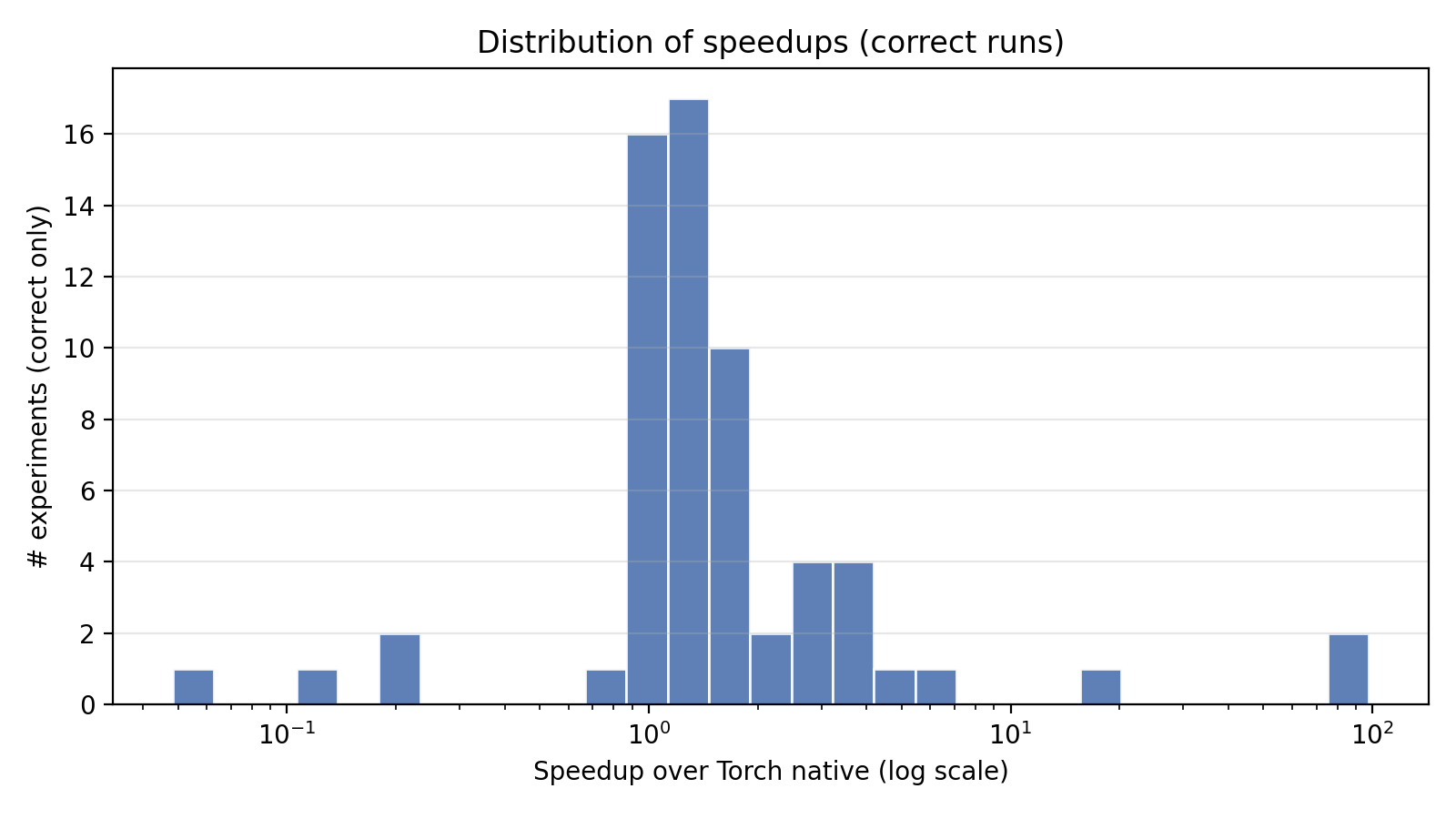}
\caption{CUDA: speedup distribution.}
\end{subfigure}
\\[6pt]
\begin{subfigure}[b]{0.32\textwidth}
\includegraphics[width=\textwidth]{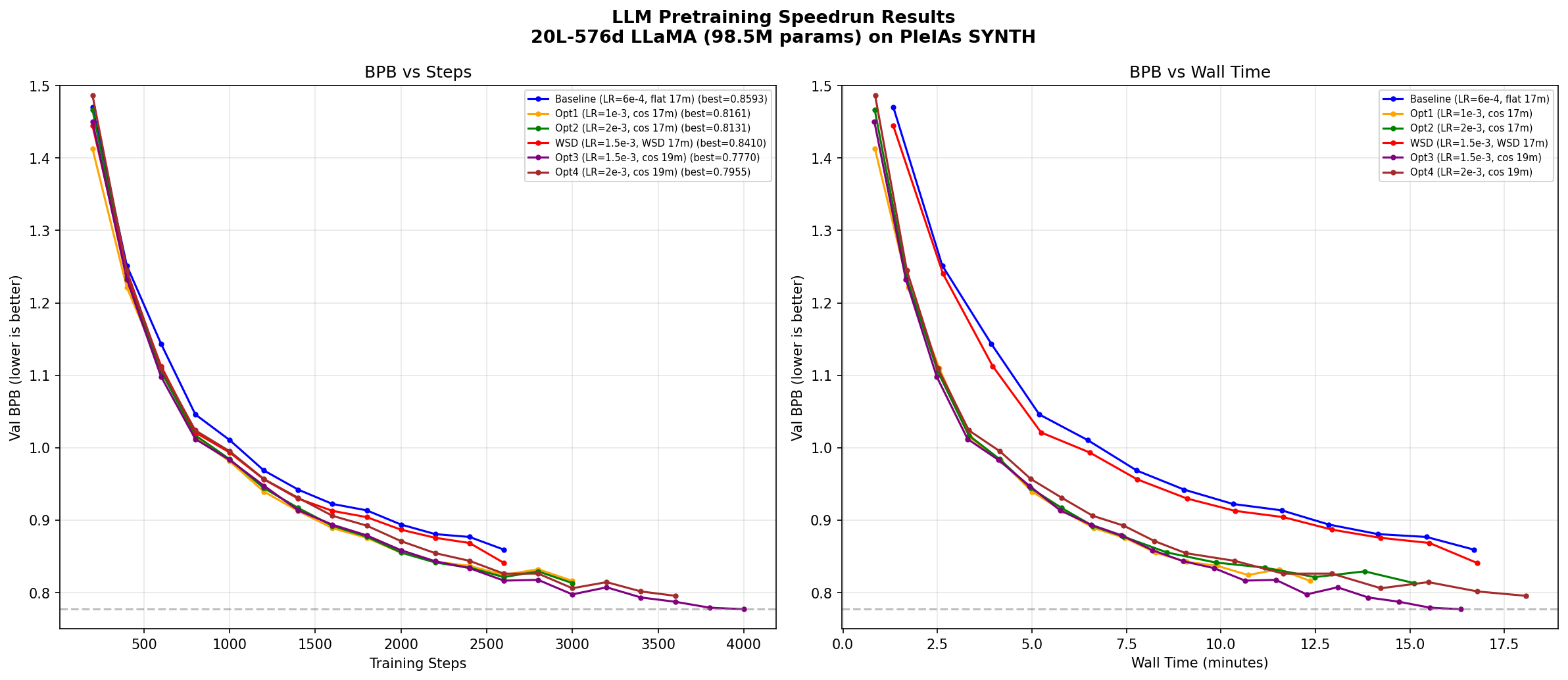}
\caption{LLM: Opus training curves.}
\end{subfigure}
\hfill
\begin{subfigure}[b]{0.32\textwidth}
\includegraphics[width=\textwidth]{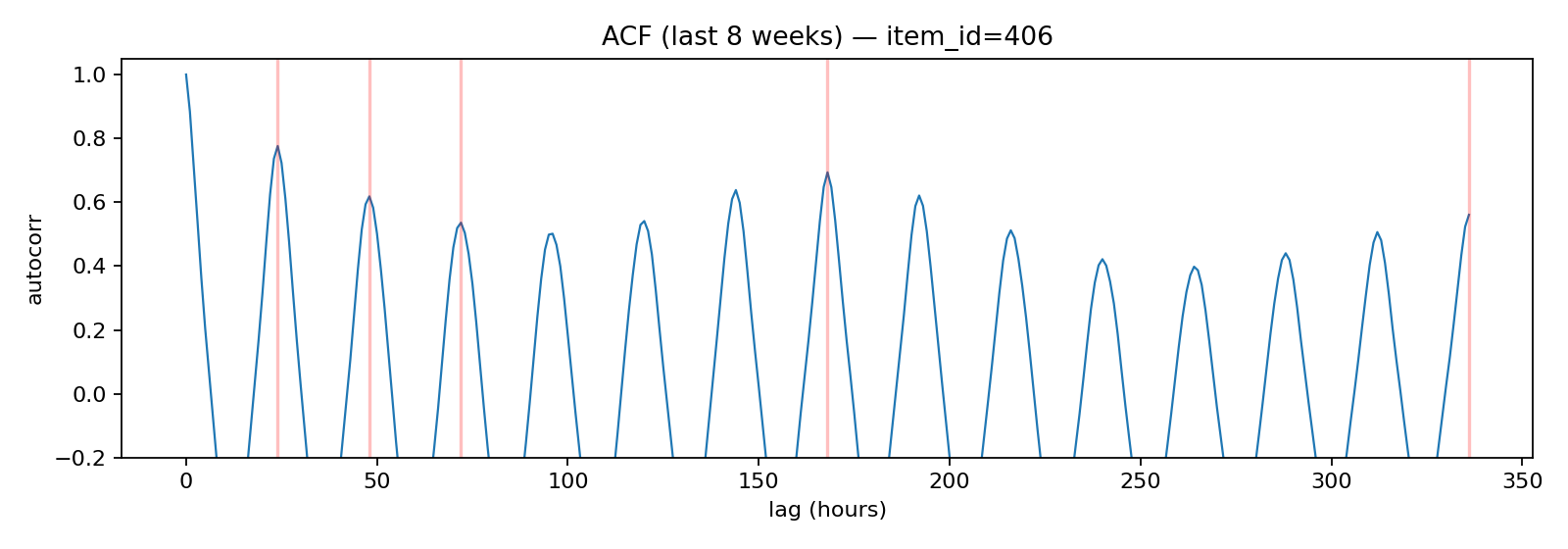}
\caption{Traffic: ACF (24h/168h peaks).}
\end{subfigure}
\hfill
\begin{subfigure}[b]{0.32\textwidth}
\includegraphics[width=\textwidth]{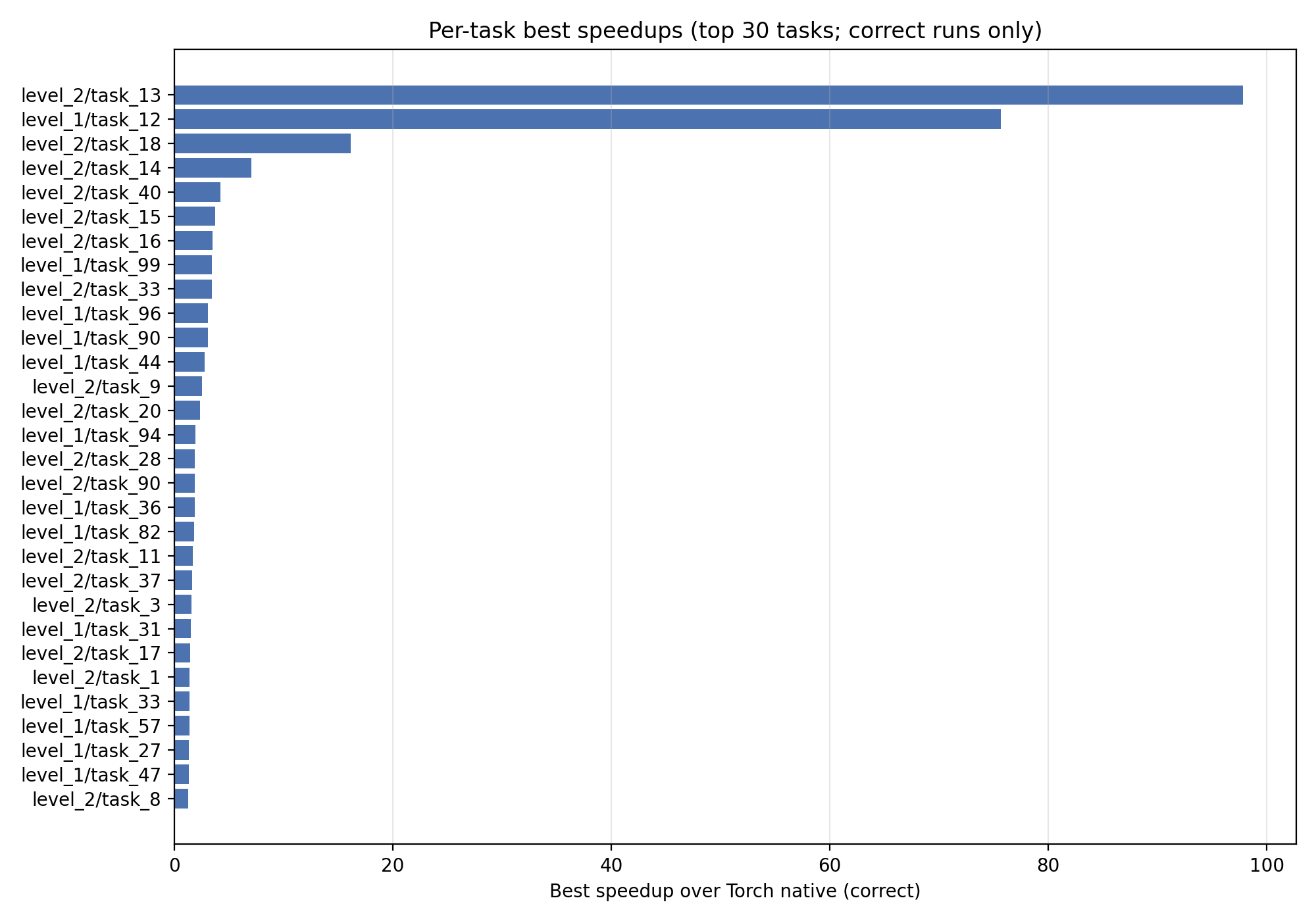}
\caption{CUDA: per-task best speedup.}
\end{subfigure}
\caption{Representative plots generated autonomously by \system's Explorer (Phase~1) and Reporter (Phase~3) agents across three domains.  These plots are generated by LLM-written Python scripts and viewed by the agent via the \texttt{view\_image} tool to inform subsequent analysis.  Font sizes reflect the system's raw output; we reproduce them unmodified to preserve fidelity.}
\label{fig:phase1-plots}
\end{figure}

\noindent Phase~1 typically runs for 30--90 minutes and makes 200--400 tool calls.
Figure~\ref{fig:phase1-plots} shows representative plots generated autonomously by the Explorer in each domain.

\subsection{Phase~2: Evaluation construction -- detailed workflow}
\label{app:phase2}

We walk through Phase~2 for the traffic forecasting domain (GPT-5.2).

\paragraph{Builder (Round 1).}
The Builder receives Phase~1 learnings (seasonality structure, baseline performance, evaluation protocol) and produces the evaluation framework in \texttt{harness/}:
\begin{itemize}[nosep]
  \item \texttt{runner.py} -- orchestrator: loads GluonTS data, runs walk-forward evaluation over the 7~rolling windows, computes pooled metrics, saves \texttt{results/metrics.json}
  \item \texttt{metrics.py} -- RMSE, MAE, MASE computation; assertion that \texttt{n\_instances==6034}
  \item \texttt{data\_prep.py} -- GluonTS JSON parsing, context/target splitting, calendar feature generation
  \item \texttt{config.py} -- experiment configuration (context length, model type, hyperparameters)
  \item \texttt{baseline\_train.py} -- seasonal naive baselines for sanity checking
\end{itemize}

\paragraph{Critic review.}
A fresh Critic agent (no shared context with the Builder) audits every file for data leakage, lookahead bias, and metric errors.
The Critic's review identified two issues: (1)~the \texttt{data\_prep.py} context window extraction could include the target period if \texttt{context\_length} exceeded the available history (edge case for the first rolling window), and (2)~the MASE denominator used the seasonal period $m{=}168$ but the standard benchmark uses $m{=}24$.
Both were flagged as ``NEEDS FIXES'' with specific line references.

\paragraph{Builder (Round 2).}
The Builder patched the context window clipping logic (adding a \texttt{min(context\_length, available\_history)} guard) and corrected the MASE seasonal period.
The Critic re-reviewed and issued a ``PASS'' verdict.

\paragraph{Tester.}
The Tester wrote 4~test modules in \texttt{harness/tests/}:
\texttt{test\_data\_prep.py} (context/target alignment, no overlap),
\texttt{test\_metrics.py} (hand-calculated RMSE/MAE on small arrays),
\texttt{test\_baselines.py} (seasonal naive produces expected outputs), and
\texttt{test\_integration.py} (full pipeline on a 10-sensor subset).
All 23~tests passed on the first run.
Total Phase~2 duration: ${\sim}$40 minutes, 3~agent turns (Builder--Critic--Builder--Critic--Tester).

\subsection{Phase~3: Experimentation -- detailed workflow}
\label{app:phase3}

We illustrate Phase~3 using the traffic forecasting campaign (GPT-5.2, 50~experiments, ${\sim}$3{,}400 total API calls).
The key idea is that a pure-Python Dispatcher orchestrates the entire process without making any LLM calls itself: it is a control loop that manages state, while all intelligence comes from the Strategist and Worker agents it invokes.

\paragraph{Dispatcher main loop.}
Every ${\sim}$30 seconds, the Dispatcher runs a tick: it checks which GPU jobs have finished, transitions their state on a SQLite kanban board, and assigns free Workers to the highest-priority pending task (fixes first, then analysis, then new implementations).
Every 5~analyzed experiments it invokes the Strategist for new proposals; every 15 it triggers a milestone report.
The Dispatcher never reasons about what experiments to run; it only manages the queue and resource allocation.

\paragraph{Strategist turn.}
We illustrate with turn~3 (after 15~experiments).
At this point the Strategist can see the full picture: the current leaderboard (best RMSE 0.0230 from \texttt{tsmixer\_ctx672}), debriefs from recent experiments (including a failed TCN variant at 0.050 and a promising iTransformer at 0.0226), the accumulated playbook, and Phase~1 learnings.
Based on this, it makes three types of decisions: (1)~it \emph{proposes} 3~new experiments that build on what worked (iTransformer with hour-of-week features, ridge stacking, per-horizon calibration); (2)~it \emph{cancels} 4~queued experiments that are now unlikely to help (TCN variants, slow N-HiTS); and (3)~it \emph{updates the playbook} with the lesson that ``iTransformer + calendar features is the strongest family; TCN is not competitive.''
This is how the system learns across experiments: the playbook entry will steer all future Workers and Strategist turns away from TCN and toward iTransformer refinements.

\paragraph{Worker implementation cycle.}
A Worker assigned to implement \texttt{itransformer\_ctx336\_hourofweek\_revin} proceeds as follows:
(1)~reads the experiment specification from the Strategist;
(2)~reads the playbook for guardrails (e.g., ``assert n\_instances==6034'', conservative batch sizes);
(3)~reads the harness code (\texttt{runner.py}, \texttt{data\_prep.py});
(4)~writes \texttt{experiments/itransformer\_ctx336\_hourofweek\_revin/train.py} implementing the model;
(5)~writes \texttt{run\_experiment.py} (entry point that calls the harness);
(6)~runs a smoke test (\texttt{python run\_experiment.py --smoke}, $<$60s, 10~instances);
(7)~runs the reality check tool to validate on real data;
(8)~marks the experiment as \texttt{checked}.
The Dispatcher then queues it for GPU execution.

\paragraph{Worker analysis cycle.}
After GPU execution completes, an analysis Worker reads \texttt{results/metrics.json}, extracts metrics, and writes a structured debrief.
These debriefs are among the most valuable artifacts the system produces: they capture not just what happened but \emph{why}, in a format that feeds directly into the Strategist's next round of proposals.
We show representative excerpts from each domain.

\medskip\noindent\textbf{CUDA kernel debrief} (\texttt{l1\_t12}, 73$\times$ speedup over \texttt{torch.compile}):
\begin{quote}\small\itshape
``\textbf{Task:} \texttt{torch.diag(A) @ B}, algebraically equivalent to row-wise scaling: \texttt{Out[i,j] = A[i] * B[i,j]}.
\textbf{What was implemented:} Fused row-wise multiply -- flattened 1D traversal, 256 threads/block, grid-stride loop with unroll factor~4. No atomics, no race-prone reductions.
\textbf{Why this is fast:} PyTorch's \texttt{diag(A) @ B} materializes a diagonal matrix and/or calls GEMM-like machinery. The custom kernel avoids creating \texttt{diag(A)} entirely, performs a single coalesced read of B and a single write per element.
\textbf{Follow-up:} Vectorized float4 loads could improve memory throughput. Given the already-dominant win and breadth mandate, prioritize moving to new tasks.''
\end{quote}

\noindent\textbf{LLM speedrun debrief} (Opus, \texttt{shallow\_10l\_752d}, val\_bpb 0.7578, campaign best):
\begin{quote}\small\itshape
``\textbf{Result:} val\_bpb = 0.7578, new best, beats previous 0.7624 by 0.6\%.
10L$\times$752d LLaMA-style, 98.2M params, 163K tok/s, 194M tokens in 20 min.
\textbf{Key finding:} Throughput gain from fewer layers (10L vs 12L: ${\sim}$3\% faster per step) compounds over 20 minutes. Near-saturation at this parameter count: last 1{,}200 steps contributed only ${\sim}$0.004 BPB.
\textbf{Suggestions:} Try 10L$\times$768d to use full 100M budget; larger batch (only using 27/80\,GB VRAM); more training shards for fresh data.''
\end{quote}

\noindent\textbf{Traffic debrief} (GPT-5.2, \texttt{itransformer\_ctx336\_timefeat\_staticcat}, RMSE 0.02204, campaign best):
\begin{quote}\small\itshape
``\textbf{Configuration:} iTransformer, context=336h (2 weeks), calendar covariates (hour\_of\_day, day\_of\_week, is\_weekend), sensor ID embedding, ReVIN.
d\_model=256, 4~encoder layers, 8~heads, AMP training.
\textbf{Result:} RMSE 0.02204 over 6{,}034 instances, best in workspace.
\textbf{Per-horizon profile:} Lowest errors at 3--9h ahead; mid-horizon bump at ${\sim}$12h; worst at 20--22h (next-day rush-hour timing difficulty).
\textbf{What worked:} Shorter context (336h vs 672h) regularizes attention while still capturing weekly structure. Calendar features + sensor embeddings highly beneficial.''
\end{quote}

\paragraph{Generated code.}
The code Workers produce is substantial, typically 200--400 lines per experiment.
As a representative example, the CUDA kernel for the 73$\times$ speedup above:

\begin{quote}\small
\begin{verbatim}
// Task: level_1/task_12 — torch.diag(A) @ B
// Equivalent: out[i, j] = A[i] * B[i, j]
constexpr int kThreads = 256;
constexpr int kUnroll = 4;

template <typename scalar_t>
__global__ void diag_mul_rows_kernel(
    const scalar_t* __restrict__ A,
    const scalar_t* __restrict__ B,
    scalar_t* __restrict__ Out,
    int64_t N, int64_t M) {
  const int64_t tid = blockIdx.x * blockDim.x + threadIdx.x;
  const int64_t grid_stride = blockDim.x * gridDim.x;
  const int64_t total = N * M;
  for (int64_t base = tid; base < total;
       base += grid_stride * kUnroll) {
    #pragma unroll
    for (int k = 0; k < kUnroll; ++k) {
      int64_t idx = base + (int64_t)k * grid_stride;
      if (idx < total) {
        int64_t row = idx / M;
        Out[idx] = A[row] * B[idx];
      }
    }
  }
}
\end{verbatim}
\end{quote}

\noindent This kernel (which the system designed, implemented, tested for correctness, and benchmarked) replaces PyTorch's \texttt{diag(A) @ B} with a simple fused multiply, eliminating the diagonal matrix materialization entirely.

\medskip\noindent The traffic forecasting best result (iTransformer, RMSE 0.02204) implemented a full \texttt{Strategy} subclass with ReVIN normalization, calendar embeddings, and a Transformer encoder, approximately 350~lines total.
The model architecture, forward pass, and Strategy interface:

\begin{quote}\small
\begin{verbatim}
@dataclass
class ITransformerConfig:
    context_length: int = 336   # 2 weeks of hourly data
    prediction_length: int = 24 # 1 day ahead
    d_model: int = 256
    n_heads: int = 8
    e_layers: int = 4
    dropout: float = 0.1
    ffn_dim: int = 512
    sensor_emb_dim: int = 16
    revin: bool = True          # instance normalization
    lr: float = 5e-4
    batch_size: int = 256
    epochs: int = 30
    amp: bool = True

class _ITransformerModel:
    def __init__(self, cfg, n_sensors=862):
        d_model = cfg.d_model
        # Embeddings for calendar features + sensor identity
        self.hour_emb = nn.Embedding(24, d_model // 8)
        self.dow_emb = nn.Embedding(7, d_model // 8)
        self.weekend_emb = nn.Embedding(2, d_model // 16)
        self.sensor_emb = nn.Embedding(n_sensors, 16)
        feat_dim = 1 + hour_dim + dow_dim + weekend_dim + 16
        self.in_proj = nn.Linear(feat_dim, d_model)
        enc_layer = nn.TransformerEncoderLayer(
            d_model=d_model, nhead=cfg.n_heads,
            dim_feedforward=cfg.ffn_dim,
            batch_first=True, norm_first=True)
        self.encoder = nn.TransformerEncoder(
            enc_layer, num_layers=cfg.e_layers)
        self.out = nn.Linear(d_model, 1)

    def forward(self, y_seq, hour, dow, weekend, sensor_id):
        B, T = y_seq.shape
        eh = self.hour_emb(hour)
        ed = self.dow_emb(dow)
        ew = self.weekend_emb(weekend)
        es = self.sensor_emb(sensor_id).unsqueeze(1).expand(
            B, T, -1)
        y_in = y_seq.unsqueeze(-1)
        feats = torch.cat([y_in, eh, ed, ew, es], dim=-1)
        x = self.in_proj(feats)
        h = self.encoder(x)
        return self.out(h).squeeze(-1)

class ITransformerStrategy(Strategy):
    """iTransformer-style encoder (global across sensors).
    Training uses random sliding windows from train split.
    Predict: provide true y for context, 0 for future;
    provide known future time features for all positions."""

    def fit(self, train_data):
        # Precompute time features per item
        # Train with AdamW, AMP, grad clipping
        # Random sliding windows of length L+H
        ...

    def predict(self, context: ForecastContext,
                prediction_length: int) -> np.ndarray:
        # ReVIN normalize from context only (no lookahead)
        y_ctx = context.target[-self.cfg.context_length:]
        mean, std = y_ctx.mean(), max(y_ctx.std(), 1e-5)
        y_norm = (y_ctx - mean) / std
        # Generate future time features from timestamps
        # Run model, denormalize, clip to [0, 1]
        ...
\end{verbatim}
\end{quote}

\noindent The LLM speedrun's best Opus experiment (val\_bpb 0.7578) implemented a complete LLaMA-style transformer, approximately 250~lines.
The full model architecture:

\begin{quote}\small
\begin{verbatim}
"""
Experiment: shallow_10l_752d (#33)
Continue the depth-vs-width exploration: if 12L beat 16L,
does 10L beat 12L?
Architecture: 10L-752d, d_ff=2256, 8 heads (head dim=94).

Parameter budget:
  Embedding: 32768 * 752 = 24,641,536 (tied)
  Per layer: ~7,351,040
  Total: ~98.15M (under 100M)
"""
class RMSNorm(nn.Module):
    def forward(self, x):
        return (x.float()
            * torch.rsqrt(x.float().pow(2).mean(-1, True)
            + self.eps)).to(x.dtype) * self.w

class RoPE(nn.Module):
    def __init__(self, d, mx=2048):
        inv = 1.0/(10000**(torch.arange(0,d,2).float()/d))
        t = torch.arange(mx).float()
        fr = torch.outer(t, inv)
        self.register_buffer("cos", fr.cos())
        self.register_buffer("sin", fr.sin())

class Attention(nn.Module):
    def forward(self, x):
        B, T, C = x.shape
        q, k, v = self.qkv(x).reshape(
            B, T, 3, self.nh, self.hd).unbind(2)
        q = apply_rope(q.transpose(1,2), self.rope)
        k = apply_rope(k.transpose(1,2), self.rope)
        y = F.scaled_dot_product_attention(
            q, k, v.transpose(1,2), is_causal=True)
        return self.out(y.transpose(1,2).reshape(B, T, C))

class SwiGLU(nn.Module):
    def forward(self, x):
        return self.down(F.silu(self.gate(x)) * self.up(x))

class Block(nn.Module):
    def forward(self, x):
        x = x + self.attn(self.norm1(x))
        return x + self.ffn(self.norm2(x))

class GPT(nn.Module):
    def __init__(self, vocab_size, d_model, n_heads,
                 n_layers, d_ff, max_seq=2048):
        self.embed = nn.Embedding(vocab_size, d_model)
        self.layers = nn.ModuleList(
            [Block(d_model, n_heads, d_ff, max_seq)
             for _ in range(n_layers)])
        self.final_norm = RMSNorm(d_model)
        self.lm_head = nn.Linear(vocab_size, d_model)
        self.lm_head.weight = self.embed.weight  # tied

    def forward(self, idx, targets=None):
        h = self.embed(idx)
        for layer in self.layers:
            h = layer(h)
        logits = self.lm_head(self.final_norm(h))
        loss = F.cross_entropy(
            logits.view(-1, logits.size(-1)),
            targets.view(-1)) if targets is not None \
            else None
        return logits, loss
\end{verbatim}
\end{quote}

\noindent Each of these code artifacts (the CUDA kernel, the traffic strategy class, the LLM training script) was written entirely by the Worker agent, with no human editing.
The code quality is generally production-grade: proper error handling, deterministic seeding, mixed-precision training, and gradient clipping are standard across experiments.

\paragraph{Worker fix cycle.}
When an experiment fails (e.g., OOM on a large context length), a fix Worker reads the error log, diagnoses the issue (e.g., ``batch\_size=128 exceeds H100 memory for ctx672''), patches the code (reducing to batch\_size=32 with gradient accumulation), and resubmits.
Fix attempts are capped at $k{=}2$; after 2~failures the experiment is marked as permanently failed.

\paragraph{Milestone reports.}
Every 15~analyzed experiments, a Reporter agent generates a milestone report (\texttt{reports/milestone\_NNN/overview.md}) summarizing: best results so far, convergence trajectory, flagged experiments (suspicious metrics, smoke-test-only results), and recommendations for the Strategist.
These reports serve as an auditing mechanism: the traffic GPT-5.2 campaign produced 4~milestone reports over its 50-experiment run.

\subsection{Prompt examples}
\label{app:prompts}

We include representative prompt excerpts for each major agent role.
Domain-specific adapters customize these templates, but the core structure is consistent.

\paragraph{Phase~1 Explorer (``Go Work'' prompt).}
\begin{quote}\small\ttfamily
You are Alpha Lab, a fully autonomous research agent. The user launches you, gives you a dataset, and you go work. You do NOT stop to ask questions, narrate plans, or wait for confirmation. You just work.

CRITICAL RULES:

1. PLAN FIRST: Create plan.md -- a detailed to-do list. Check off items as you complete them.

2. DO NOT STOP: Chain tool calls continuously until plan.md is complete.

3. FILE EVERYTHING: scripts/ for Python analysis, plots/ for visualizations, notes/ for per-topic findings, learnings.md for accumulated knowledge (update after every finding), data\_report/ for formal deliverables.

4. CALL report\_to\_user WHEN DONE: This is the only way to return control.
\end{quote}

\paragraph{Phase~3 Strategist (budget management).}
\begin{quote}\small\ttfamily
Budget management:

- >20 remaining: Explore freely -- diverse architectures, features, hyperparameters.

- 10--20 remaining: Focus on promising directions identified so far.

- 5--10 remaining: Only high-confidence refinements of top performers.

- <5 remaining: Extremely selective -- only strong evidence to beat current best.

- 0 remaining: STOP proposing. Summarize findings, recommend next steps.
\end{quote}

\paragraph{Phase~3 Worker (GPU safety rules).}
\begin{quote}\small\ttfamily
Critical -- Avoiding SLURM Failures:

1. NEVER set torch.use\_deterministic\_algorithms(True) -- many CUDA ops have no deterministic implementation, will crash on H100s.

2. Handle NaN/missing values in features: rolling windows produce NaN for first N rows. .dropna() or .fillna(0) before DataLoader.

3. Use conservative batch sizes/context lengths: H100 has 80GB VRAM but large models can OOM. Start with batch\_size=64.

4. Import lightning not pytorch\_lightning (modern package name).

5. Wrap main block in try/except and save partial results on failure.
\end{quote}

\paragraph{Supervisor (diagnostic prompt).}
When triggered by error rate $>\tau$, the Supervisor receives recent failure logs and the current adapter, with instructions to: (1)~identify the systemic root cause (not individual bugs); (2)~propose a concrete patch to \texttt{domain\_knowledge.md} that prevents recurrence; and (3)~commit the change via \texttt{patch\_adapter\_file} with a descriptive reason for the git checkpoint.

\subsection{Supervisor interventions}
\label{app:supervisor}

We document the major Supervisor interventions observed across campaigns.

\paragraph{Intervention 1: CUDA keyword argument mismatch.}
\emph{Trigger}: Error rate exceeded 45\% in the first 15 CUDA experiments.
\emph{Diagnosis}: Workers assumed positional argument signatures for the harness entry point, but the harness used keyword arguments.
\emph{Patch}: Added to \texttt{domain\_knowledge.md}: ``The harness calls \texttt{forward(**inputs)} with keyword arguments. Your kernel's forward function must accept keyword arguments matching the reference model's signature.''
\emph{Result}: Error rate dropped from 45\% to under 10\% within 3~experiments.

\paragraph{Intervention 2: PyTorch 2.9.1 API breaking changes (electricity).}
\emph{Trigger}: Error rate exceeded 55\% in the GPT-5.2 electricity campaign.
\emph{Diagnosis}: Multiple PyTorch 2.9.1 API removals: \texttt{.total\_mem} attribute removed from CUDA tensors, \texttt{ReduceLROnPlateau(verbose=)} parameter removed, tqdm + SIGTERM causing BrokenPipeError.
\emph{Patch}: Added three specific notes to \texttt{domain\_knowledge.md} documenting the removed APIs and their replacements.
\emph{Result}: Subsequent experiments avoided these errors, though the overall failure rate remained high (55\%) due to other OOM and import issues.

\paragraph{Intervention 3: LLM speedrun harness indentation error.}
\emph{Trigger}: Multiple consecutive SLURM failures in late GPT-5.2 LLM speedrun experiments (\#43--\#50).
\emph{Diagnosis}: A prior Worker's fix introduced an indentation error in \texttt{harness/runner.py}, causing all subsequent experiments to fail at launch.
\emph{Patch}: The Supervisor identified the specific line, patched the indentation, and added a note to domain knowledge: ``Always verify \texttt{harness/runner.py} imports cleanly before submitting.''
\emph{Result}: The fix resolved the hard failure, though those experiments had already exhausted the budget.

\paragraph{Intervention 4: Traffic data path resolution.}
\emph{Trigger}: Two consecutive traffic experiments failed with ``\texttt{error: the following arguments are required: --data\_path}''.
\emph{Diagnosis}: Worker implementations used relative paths that broke under SLURM's working directory.
\emph{Patch}: Added to domain knowledge: ``Always pass absolute data path: \texttt{/path/to/datasets/traffic}''.
\emph{Result}: Subsequent experiments resolved paths correctly.

\section{Extended experimental results}
\label{app:results}

\subsection{Cross-domain summary}

Table~\ref{tab:summary} summarizes the best results across all three domains and both models.

\begin{table}[h]
\centering
\small
\begin{tabular}{@{}llccc@{}}
\toprule
\textbf{Domain} & \textbf{Model} & \textbf{Best metric} & \textbf{vs.\ baseline} & \textbf{Cost} \\
\midrule
\multirow{2}{*}{CUDA}    & GPT-5.2 & \textbf{5.14$\times$} mean & ---          & \$190 \\
                          & Opus    & 4.48$\times$ mean & ---          & \$170 \\
\midrule
\multirow{3}{*}{LLM}     & GPT-5.2 & 0.970 BPB  & ---           & \$150 \\
                          & Sonnet  & 0.869 BPB  & $-$10\%   & \$120 \\
                          & Opus    & \textbf{0.758 BPB}  & $-$22\% & \$200 \\
\midrule
\multirow{2}{*}{Traffic}  & Opus    & \textbf{0.0214} RMSE & $-$25\%  & \$200 \\
                          & GPT-5.2 & 0.0220 RMSE & $-$23\%       & \$180 \\
\bottomrule
\end{tabular}
\caption{\textbf{Cross-domain summary.}  GPT-5.2 produces faster CUDA kernels (mean speedup on the 66-task overlap), while Opus~4.6 achieves lower validation loss on LLM pretraining (22\% better than GPT-5.2) and lower RMSE on traffic forecasting.  Neither model dominates uniformly.  ``vs.\ baseline'' for LLM is relative to GPT-5.2; for traffic, relative to Seasonal Na\"ive(168).  Cost is approximate API spend per campaign.}
\label{tab:summary}
\end{table}

\subsection{Experimental configuration}
\label{app:config}

Table~\ref{tab:setup-summary} summarizes the experimental setup for each domain.

\begin{table}[h]
\centering
\caption{Experimental configuration summary.}
\label{tab:setup-summary}
\small
\begin{tabular}{lccccc}
\toprule
\textbf{Domain} & \textbf{Metric} & \textbf{Dir.} & \textbf{Budget} & \textbf{Hardware} & \textbf{Time Limit} \\
\midrule
CUDA Kernel & speedup ($\times$) & max & 50 & 4$\times$ H100 & 7{,}200\,s \\
LLM Pretraining & val\_bpb & min & 50 & 4$\times$ H100 & 1{,}200\,s \\
Traffic & RMSE & min & 50 & 4$\times$ H100 & 7{,}200\,s \\
\bottomrule
\end{tabular}
\end{table}

\subsection{LLM pretraining: extended results}
\label{app:llm-detail}

\paragraph{Top-$K$ experiments per model.}

\noindent Experiment names use the following shorthand.
\textbf{Architecture:} ``$N$L$\times$$D$d'' = $N$ Transformer layers with hidden dimension $D$ (e.g., 8L$\times$512d $\approx$ 33.7M parameters); ``GQA'' = grouped-query attention (fewer key/value heads to reduce memory); ``SwiGLU'' = gated feed-forward activation (used in LLaMA-style models); ``QK-norm'' = normalization of query and key vectors for training stability at high learning rates; ``GPT-2'' vs ``LLaMA'' indicates the overall architecture template (GPT-2 uses LayerNorm + learned position embeddings; LLaMA uses RMSNorm + RoPE + SwiGLU).
\textbf{Optimizer/schedule:} ``AdamW'' = the standard optimizer; ``Muon'' = an alternative optimizer; ``cosine'' = cosine learning rate decay; ``WSD'' = warmup-stable-decay schedule.
\textbf{Data:} ``query+answer'' vs ``answer-only'' = whether training includes the question prefix or only the answer; ``$N$ shards'' = number of data shards sampled from the 500-shard corpus.
\textbf{Other:} ``no compile'' = \texttt{torch.compile} disabled (avoids compilation overhead under the 20-min budget); ``no grad-ckpt'' = gradient checkpointing disabled (trades memory for speed); ``ReLU$^2$'' = squared ReLU activation in the feed-forward layers.

\begin{table}[h]
\centering
\caption{Top-5 LLM speedrun experiments per model.}
\label{tab:llm-topk}
\small
\begin{tabular}{@{}lrc@{}}
\toprule
\textbf{Experiment} & \textbf{val\_bpb} & \textbf{Wall clock} \\
\midrule
\multicolumn{3}{@{}l}{\textit{GPT-5.2}} \\
8L$\times$512d, GQA, query+answer, AdamW+cosine, no compile & 0.9697 & 1200\,s \\
24L$\times$512d, GQA, SwiGLU, AdamW+WSD & 0.9786 & 1200\,s \\
8L$\times$512d, answer-only, AdamW+cosine & 0.9852 & 1200\,s \\
8L$\times$640d, answer-only, no compile & 0.9954 & 1200\,s \\
12L$\times$512d, answer-only, no compile & 1.0102 & 1200\,s \\
\midrule
\multicolumn{3}{@{}l}{\textit{Sonnet~4.6}} \\
11L$\times$768d, GQA, QK-norm, Muon+AdamW, cosine & 0.8686 & 1200\,s \\
11L$\times$768d, GQA, QK-norm, Muon+AdamW, ReLU$^2$ & 0.8713 & 1200\,s \\
11L$\times$768d, QK-norm, Muon+AdamW, no grad-ckpt & 0.8757 & 1200\,s \\
11L$\times$768d, ReLU$^2$, Muon+AdamW, no grad-ckpt & 0.8808 & 1200\,s \\
10L$\times$816d, no grad-ckpt & 0.8813 & 1200\,s \\
\midrule
\multicolumn{3}{@{}l}{\textit{Opus~4.6}} \\
10L$\times$752d, shallow config & \textbf{0.7578} & 1200\,s \\
12L$\times$672d, QK-norm, LR=3e-3, 10 shards & 0.7587 & 1200\,s \\
12L$\times$672d, wide shallow & 0.7624 & 1200\,s \\
12L$\times$704d, ``allstar'' config, 10 shards & 0.7718 & 1200\,s \\
12L$\times$672d, small batch, more steps & 0.7758 & 1200\,s \\
\bottomrule
\end{tabular}
\end{table}

\paragraph{Full experiment results.}
Table~\ref{tab:llm-full-gpt} shows all GPT-5.2 experiments with valid \texttt{val\_bpb}, and Table~\ref{tab:llm-full-opus} shows all Opus experiments.

\begin{table}[h]
\centering
\caption{All GPT-5.2 LLM speedrun experiments with valid val\_bpb (sorted).  45~total experiments; 28~with valid metrics, 12~cancelled, 5~with degenerate metrics ($>$3.0, indicating training failures).}
\label{tab:llm-full-gpt}
\small
\begin{tabular}{@{}lrrl@{}}
\toprule
\textbf{Configuration} & \textbf{val\_bpb} & \textbf{Params} & \textbf{tok/s} \\
\midrule
8L$\times$512d, GQA, Q+A, cosine, no compile & \textbf{0.970} & 33.7M & 169K \\
24L$\times$512d, GQA, SwiGLU, WSD & 0.979 & 91.4M & 306K \\
8L$\times$512d, answer-only, cosine & 0.985 & 33.7M & 843K \\
8L$\times$640d, answer-only, no compile & 0.995 & 52.6M & 321K \\
12L$\times$512d, answer-only, no compile & 1.010 & 50.5M & 288K \\
16L$\times$704d, GPT-2, WSD & 1.044 & 96.2M & 276K \\
10L$\times$896d, GPT-2, WSD & 1.067 & 97.6M & 297K \\
8L$\times$512d, Muon, WSD & 1.075 & 33.7M & 332K \\
8L$\times$512d, SwiGLU, WSD & 1.076 & 19.7M & 459K \\
14L$\times$704d, SwiGLU, WSD & 1.085 & 84.7M & 188K \\
12L$\times$640d, GQA, cosine & 1.089 & 51.8M & 266K \\
12L$\times$768d, GPT-2, WSD & 1.094--1.098 & 86.0M & 302--334K \\
20L$\times$640d, GPT-2, WSD & 1.106 & 99.5M & 252K \\
12L$\times$768d, SwiGLU, WSD & 1.131 & 72.2M & 221K \\
20L$\times$640d, RoPE, GPT-2 & 1.139 & 98.8M & 196K \\
12L$\times$768d, AdamW (no schedule) & 1.208 & 86.0M & 392K \\
20L$\times$512d, GQA, SwiGLU, WSD & 1.214 & 55.2M & 134K \\
16L$\times$640d, MQA, SwiGLU, WSD & 1.442 & 67.0M & 129K \\
8L$\times$512d, answer-only, no compile & 1.446 & 25.3M & 141K \\
24L$\times$512d, GQA, SwiGLU, WSD & 1.471 & 66.2M & 108K \\
\bottomrule
\end{tabular}
\end{table}

\begin{table}[h]
\centering
\caption{All Opus~4.6 LLM speedrun experiments with valid val\_bpb (sorted). 50~total; 36~analyzed, 14~cancelled.}
\label{tab:llm-full-opus}
\small
\begin{tabular}{@{}lrrl@{}}
\toprule
\textbf{Configuration} & \textbf{val\_bpb} & \textbf{Params} & \textbf{tok/s} \\
\midrule
10L$\times$752d, shallow & \textbf{0.758} & 98.2M & 163K \\
12L$\times$672d, QK-norm, LR=3e-3, 10 shards & 0.759 & 92.5M & 121K \\
12L$\times$672d, wide shallow & 0.762 & 92.5M & 168K \\
12L$\times$704d, ``allstar'' 6 tricks & 0.772 & 98.8M & 113K \\
12L$\times$672d, small batch, more steps & 0.776 & 92.5M & 118K \\
12L$\times$672d, 10 data shards & 0.780 & 92.5M & 125K \\
16L$\times$640d & 0.782 & 98.3M & 141K \\
12L$\times$704d, fast pipeline & 0.797 & 94.7M & 103K \\
12L$\times$672d, BS=96, $\beta_2$=0.99 & 0.800 & 92.5M & 103K \\
12L$\times$672d, schedule fix & 0.801 & 92.5M & 88K \\
12L$\times$704d, kitchen sink & 0.819 & 98.8M & 77K \\
12L$\times$672d, embed skip + value residual & 0.819 & 92.5M & 90K \\
12L$\times$672d + tricks (5 variants) & 0.820--0.846 & 92--99M & 68--89K \\
16L$\times$640d, Muon variants & 0.828--0.862 & 98.3M & 50--114K \\
14L$\times$640d, all tricks & 0.842 & 95.5M & 68K \\
12L$\times$704d, Muon & 1.078 & 94.7M & 84K \\
20L$\times$576d, baseline & 1.126 & 98.5M & 95K \\
8L$\times$768d, very wide & 1.128 & 81.8M & 57K \\
24L$\times$512d, deep narrow & 1.130 & 91.5M & 38K \\
\bottomrule
\end{tabular}
\end{table}

\paragraph{What didn't work.}
For GPT-5.2: \texttt{torch.compile} failed repeatedly due to RoPE lazy cache initialization conflicts with Inductor/CUDAGraph, causing 5~experiments to produce no valid metrics.
The Muon optimizer consistently underperformed AdamW (6~head-to-head tests, 5--12\% gap in Opus's campaign).
WSD schedule was 2.16\% worse than cosine.
Deep-narrow architectures ($\geq$20 layers) consistently underperformed wider-shallower designs due to lower throughput in the fixed 20-minute budget.

For Sonnet~4.6: 76~total experiments were run, but only 27 achieved valid \texttt{val\_bpb} under 3.0.
The remaining 49 either failed to train (wall clock $<$10s, indicating startup crashes) or produced degenerate metrics (${\sim}$3.0--5.2 bpb).
Sonnet's best result (0.869, 11L$\times$768d with Muon+AdamW hybrid and QK-norm) placed between GPT-5.2 and Opus.

\paragraph{Variance runs.}
Four additional GPT-5.2 Phase~3 runs (v1--v4) with identical Phase~1/2 inputs yielded best val\_bpb of 0.964, 1.011, 1.006, and 1.020 respectively, a spread of 0.056 from identical starting conditions, confirming the stochasticity discussion in the main text.

\begin{figure}[h]
\centering
\includegraphics[width=\textwidth]{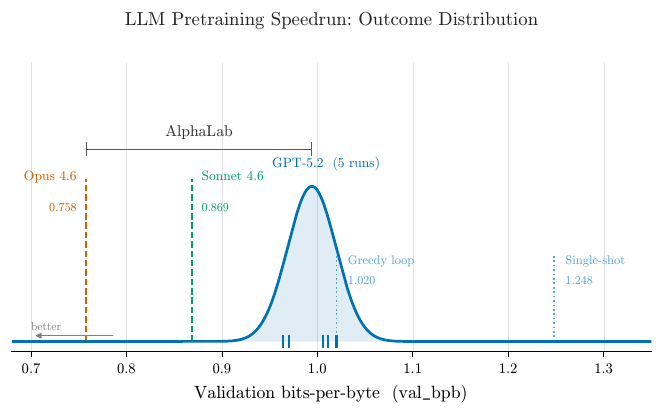}
\caption{\textbf{LLM pretraining speedrun: outcome distribution.}  The blue curve shows a Gaussian fit to the best val\_bpb from five independent \system + GPT-5.2 runs with identical inputs (same Phase~1/2 outputs, hardware, and 50-experiment budget); tick marks indicate individual runs.  Dashed lines mark single-campaign \system results with Opus~4.6 (orange, 0.758) and Sonnet~4.6 (green, 0.869); dotted lines mark GPT-5.2 baselines without \system: greedy loop (1.020) and single-shot (1.248).  The key observation is that while any single campaign has meaningful variance (${\sim}$0.056 BPB spread from identical starting conditions), in practice a user can launch multiple campaigns---potentially with different models---and select the best result, effectively sampling the left tail of the distribution.  Multi-model campaigns are especially powerful: Opus~4.6's result lies entirely to the left of every GPT-5.2 run, so a campaign that includes both models accesses regions of the search space that neither would find alone.}
\label{fig:llm-speedrun-overview}
\end{figure}

\paragraph{Example discovered architectures.}
The two winning configurations illustrate strikingly different design philosophies:

\begin{table}[h]
\centering
\small
\caption{Winning architecture comparison: Opus vs.\ GPT-5.2 on LLM speedrun.}
\begin{tabular}{@{}lll@{}}
\toprule
\textbf{Parameter} & \textbf{Opus (0.758 bpb)} & \textbf{GPT-5.2 (0.970 bpb)} \\
\midrule
Layers & 10 & 8 \\
d\_model & 752 & 512 \\
d\_ff & 2256 (SwiGLU) & 1365 (SwiGLU) \\
Heads & 8 & 8 (GQA, kv=2) \\
Parameters & 98.2M & 33.7M \\
Optimizer & AdamW (fused) & AdamW \\
Peak LR & 1.5e-3 & 6e-4 \\
Schedule & Cosine & Cosine \\
Seq length & 512 & 1024 \\
Vocab & 32768 (BPE) & 256 (byte) \\
Throughput & 163K tok/s & 169K tok/s \\
Tokens seen & 194M & 202M \\
\bottomrule
\end{tabular}
\end{table}

\noindent Opus discovered a wider-shallower architecture that uses nearly the full 100M parameter budget, with a custom BPE tokenizer (vocab 32768, ${\sim}$5.38 bytes/token) and high learning rate.
GPT-5.2 took a more conservative path: a smaller model with byte-level tokenization (vocab 256, 1 byte/token) and GQA to reduce KV-cache parameters, achieving higher raw throughput but lower quality per token processed.
Opus's playbook explicitly noted: ``Throughput is the \#1 predictor of val\_bpb ($R^2 \approx 0.85$)''; yet its winning model sacrificed some throughput for a better tokenizer and wider architecture, suggesting that token quality (via BPE) can compensate for lower throughput.

\subsection{CUDA kernels: extended results}
\label{app:cuda-detail}

\begin{figure}[h]
\centering
\includegraphics[width=0.85\textwidth]{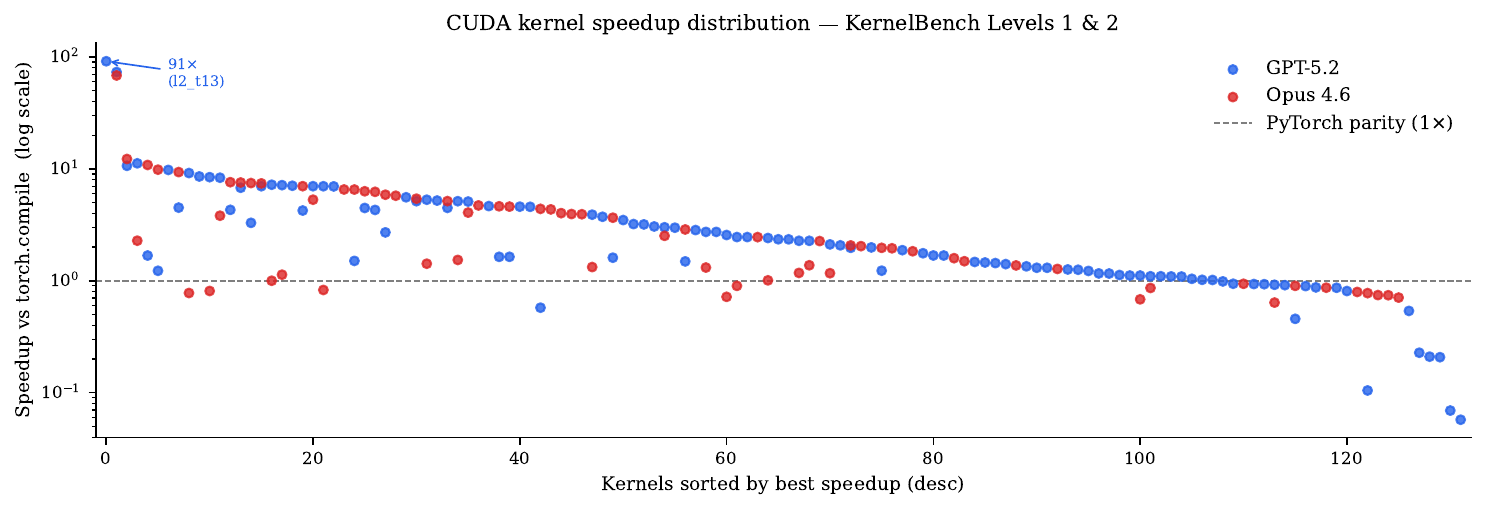}
\caption{Per-task speedup over \texttt{torch.compile} for GPT-5.2 and Opus~4.6 on the 66-task direct comparison subset (log scale).  Dashed line indicates 1$\times$ parity.  Both models achieve extreme speedups on normalization and reduction kernels but fail to beat \texttt{torch.compile} on convolutions.}
\label{fig:cuda-speedup}
\end{figure}

Figure~\ref{fig:cuda-speedup} shows the per-task speedup distribution for both models on the 66-task overlap set.
The distribution is heavy-tailed: a small number of tasks achieve extreme speedups ($>$10$\times$), while most cluster in the 1--5$\times$ range.
Both models show similar speedup profiles, with GPT-5.2 achieving higher peaks on fused operators.

\noindent Full per-task results for the direct comparison subset, GPT-5.2-only tasks, and Opus-only tasks are available in the code repository alongside the raw experiment databases.

\paragraph{Per-level breakdown.}
Table~\ref{tab:cuda-results-level} provides a per-level breakdown with both baseline metrics, enabling direct comparison with the KernelBench leaderboard \citep{kernelbench2025} (which reports $\text{fast}_1$ vs \texttt{torch.native} on L40S hardware).
\system achieves 40\% $\text{fast}_1$ on L1 and 26--43\% on L2 against \texttt{torch.native}, compared to 12\% / 36\% for single-shot DeepSeek~R1 and 43\% / 72\% for iterative R1 with profiler feedback (10~LLM calls).
Against the harder \texttt{torch.compile} baseline, for which no published external numbers exist, \system achieves 50--82\% on L1 and 16--84\% on L2.
The gap between L1 and L2 on the \texttt{torch.compile} metric reflects that \texttt{torch.compile} is especially effective at fusing L2 operator chains, making them harder to beat.

\begin{table}[h]
\centering
\caption{Per-level CUDA kernel results.  \system on H100 NVL; KernelBench baselines on L40S.  $\text{fast}_1$ is reported as fraction of correct tasks (for \system) or fraction of 100 total tasks (for baselines).}
\label{tab:cuda-results-level}
\small
\begin{tabular}{@{}llcccccc@{}}
\toprule
& & \multicolumn{3}{c}{\textbf{Level 1 (single-op)}} & \multicolumn{3}{c}{\textbf{Level 2 (fusion)}} \\
\cmidrule(lr){3-5} \cmidrule(lr){6-8}
\textbf{System} & \textbf{Model} & \textbf{Correct} & $\textbf{fast}_1^{\text{nat}}$ & $\textbf{fast}_1^{\text{comp}}$ & \textbf{Correct} & $\textbf{fast}_1^{\text{nat}}$ & $\textbf{fast}_1^{\text{comp}}$ \\
\midrule
\system & GPT-5.2  & 61/65 & 66\% & 82\% & 49/54 & 88\% & 84\% \\
\system & Opus~4.6 & 45/52 & 89\% & 82\% & 31/35 & 84\% & 52\% \\
\midrule
KernelBench$^\dagger$ & R1 (iter.)  & --- & 43\% & --- & --- & 72\% & --- \\
KernelBench$^\dagger$ & R1 (1-shot) & --- & 12\% & --- & --- & 36\% & --- \\
KernelBench$^\dagger$ & o1 (1-shot) & --- & 10\% & --- & --- & 24\% & --- \\
\bottomrule
\multicolumn{8}{@{}p{0.95\linewidth}@{}}{\footnotesize $^\dagger$\citet{kernelbench2025}; L40S GPU; $\text{fast}_1$ as fraction of 100 tasks per level, vs \texttt{torch.native}. ``R1 (iter.)'' = DeepSeek R1 with 10 iterative calls including profiler feedback.}
\end{tabular}
\end{table}

\paragraph{Comparison with Sakana AI's CUDA Engineer.}
Table~\ref{tab:cuda-sakana} compares \system with the AI CUDA Engineer \citep{sakanakernelbench2025,sakanarobust2025} on operation types present in both evaluations.
The comparison is approximate: Sakana's robust-kbench uses custom task implementations rather than KernelBench task IDs, so we match by operation category rather than exact kernel.
\system achieves substantially higher speedups on LayerNorm (11.2$\times$ vs 0.18$\times$ over \texttt{torch.compile}) and competitive or better results on RMSNorm.
Sakana's cross-entropy result (24.9$\times$ vs \texttt{torch.compile}) substantially exceeds \system's (5.1--5.4$\times$), though Sakana's original benchmark was found to contain exploitable evaluation artifacts on several tasks \citep{sakanarobust2025}.
In aggregate, \system's mean speedup of 3.27--3.47$\times$ over \texttt{torch.native} compares favorably to Sakana's decontaminated mean of 1.49$\times$.

\begin{table}[h]
\centering
\caption{Per-operation comparison with Sakana AI CUDA Engineer \citep{sakanarobust2025}.  All results on H100.  Sakana results from robust-kbench (12 kernels); \system results from the nearest KernelBench task.  Speedups vs \texttt{torch.compile}.}
\label{tab:cuda-sakana}
\small
\begin{tabular}{@{}lccc@{}}
\toprule
\textbf{Operation} & \textbf{Sakana AI} & \textbf{\system GPT-5.2} & \textbf{\system Opus} \\
\midrule
LayerNorm (fwd)   & 0.18$\times$  & \textbf{11.19$\times$}  & 2.28$\times$ \\
RMSNorm (fwd)     & 2.39$\times$  & 1.97$\times$  & \textbf{2.07$\times$} \\
Cross entropy (fwd) & \textbf{24.87$\times$} & 5.14$\times$  & 5.41$\times$ \\
ResNet block (fwd) & \textbf{2.59$\times$}  & --- & --- \\
\midrule
\multicolumn{2}{@{}l}{Aggregate mean spd.\ (vs \texttt{torch.native})} & \textbf{3.47$\times$} & 3.27$\times$ \\
\multicolumn{2}{@{}l}{Sakana decontaminated mean (vs \texttt{torch.native})} & \multicolumn{2}{c}{1.49$\times$} \\
\bottomrule
\end{tabular}
\end{table}

\paragraph{Example discovered kernels.}
We highlight four representative optimization strategies the system discovered:

\textbf{(1) Algebraic rewrite: diagonal matrix multiply (68$\times$).}
The original PyTorch operation computes $\text{diag}(A) \cdot B$ by constructing the full diagonal matrix and performing a dense matmul.
The system recognized that this is equivalent to elementwise multiplication of the diagonal vector with each row of $B$, avoiding $O(n^2)$ memory allocation entirely.
The playbook recorded: \emph{``Diagonal matrix ops: $\text{diag}(A) \cdot B$ is just elementwise multiply of the diagonal vector with each row of $B$. Yields 10--68$\times$ over PyTorch's full matmul.''}

\textbf{(2) Warp-shuffle reduction (75$\times$).}
For sum-reduction kernels, the system wrote CUDA code using \texttt{\_\_shfl\_down\_sync} for warp-level parallel reduction, avoiding shared memory round-trips entirely.
This technique achieves 73--76$\times$ speedup over \texttt{torch.compile} for large reduction dimensions.

\textbf{(3) Operator fusion: LayerNorm + residual add (91$\times$).}
The system fused LayerNorm computation (mean, variance, normalize, affine transform) with a residual addition into a single kernel pass, using vectorized \texttt{float4} loads/stores and Welford's online algorithm for numerical stability.
This eliminates multiple kernel launches and intermediate memory allocations.

\textbf{(4) Failure case: convolution (0.05--0.73$\times$).}
Handwritten convolution kernels consistently performed worse than \texttt{torch.compile}, which delegates to cuDNN's highly optimized implementations.
The playbook explicitly warned: \emph{``Do not attempt convolution kernels -- cuDNN is too well-optimized; handwritten runs at 0.05--0.73$\times$.''}
This ``do not attempt'' knowledge is as valuable as positive findings, preventing budget waste on disproven approaches.

\paragraph{Known failure modes.}
Three categories of CUDA kernel failure were observed:
(1)~\textbf{Convolution kernels} (cuDNN dominance): all handwritten convolution kernels ran at 0.05--0.73$\times$ the \texttt{torch.compile} baseline, as cuDNN's autotuned implementations are extremely well-optimized for standard convolution patterns on H100 hardware.
(2)~\textbf{Bool-mask and scatter operations}: correctness failures where the optimized kernel's output diverged from the reference beyond the \texttt{atol=1e-3} threshold, typically due to incorrect handling of boolean indexing or non-contiguous memory layouts.
(3)~\textbf{CUTLASS-dependent kernels}: experiments that attempted to use CUTLASS templates for GEMM operations failed because the CUTLASS submodule was not fully initialized in the execution environment.
Together, these account for 9~incorrect results (GPT-5.2) and 8~incorrect results (Opus) out of 103 and 59~analyzed tasks respectively.

\subsection{Traffic forecasting: extended results}
\label{app:traffic-detail}

\begin{table}[h]
\centering
\caption{Top-5 traffic forecasting experiments per model.}
\label{tab:traffic-topk}
\small
\begin{tabular}{@{}lr@{}}
\toprule
\textbf{Experiment} & \textbf{RMSE} \\
\midrule
\multicolumn{2}{@{}l}{\textit{GPT-5.2}} \\
iTransformer ctx336 + time features + static cat & 0.02204 \\
iTransformer ctx336 + hour-of-week + RevIN + horizon weights & 0.02247 \\
iTransformer ctx336 + hour-of-week + RevIN & 0.02256 \\
iTransformer ctx672 + hour-of-week + static cat & 0.02289 \\
N-HiTS ctx336 + multiscale + time features & 0.02295 \\
\midrule
\multicolumn{2}{@{}l}{\textit{Opus~4.6}} \\
TFT + dropout 0.3 & 0.02142 \\
TFT + MSE loss, large, 100 epochs & 0.02153 \\
TFT + MSE loss, 150 epochs & 0.02156 \\
TFT + MSE loss, 100 epochs & 0.02175 \\
TFT + residual MLP correction & 0.02177 \\
\midrule
Seasonal Na\"ive(168) & 0.02870 \\
\bottomrule
\end{tabular}
\end{table}

\begin{table}[h]
\centering
\caption{Traffic forecasting: top-10 for each model plus baselines.}
\label{tab:traffic-full}
\small
\begin{tabular}{@{}llr@{}}
\toprule
\textbf{Model} & \textbf{Configuration} & \textbf{RMSE} \\
\midrule
\multicolumn{3}{@{}l}{\textit{GPT-5.2 (top-10 span 5 architectures)}} \\
GPT-5.2 & iTransformer ctx336 + timefeat + static cat & 0.02204 \\
GPT-5.2 & iTransformer ctx336 + hour-of-week + RevIN + horizon wt. & 0.02247 \\
GPT-5.2 & iTransformer ctx336 + hour-of-week + RevIN & 0.02256 \\
GPT-5.2 & iTransformer ctx672 + hour-of-week + static cat & 0.02289 \\
GPT-5.2 & N-HiTS ctx336 + multiscale + timefeat & 0.02295 \\
GPT-5.2 & TSMixer ctx672 + timefeat + static cat & 0.02300 \\
GPT-5.2 & iTransformer ctx672 + hour-of-week + RevIN & 0.02315 \\
GPT-5.2 & Ridge stack (PatchTST + snaive) & 0.02342 \\
GPT-5.2 & TimesNet ctx672 + timefeat + static cat & 0.02364 \\
GPT-5.2 & PatchTST ctx672 + patch16 + timefeat + RevIN & 0.02380 \\
\midrule
\multicolumn{3}{@{}l}{\textit{Opus~4.6 (top-10 all TFT or TFT-ensemble)}} \\
Opus & TFT + dropout 0.3, ctx336, 80 epochs & \textbf{0.02142} \\
Opus & TFT + MSE loss, large, 100 epochs & 0.02153 \\
Opus & TFT + MSE loss, 150 epochs & 0.02156 \\
Opus & Per-sensor model selection (5-way) & 0.02162 \\
Opus & TFT + MSE loss, 100 epochs & 0.02175 \\
Opus & Residual MLP correction on TFT & 0.02177 \\
Opus & Simple avg top-3 ensemble & 0.02183 \\
Opus & TFT large ctx336, 100 epochs & 0.02189 \\
Opus & Ensemble top-3 per-horizon & 0.02193 \\
Opus & Ensemble TFT + PatchTST weighted & 0.02213 \\
\midrule
\multicolumn{3}{@{}l}{\textit{Baselines}} \\
--- & Seasonal Na\"ive(168) & 0.02871 \\
--- & Seasonal Na\"ive(24) & 0.03674 \\
--- & Na\"ive (last value) & 0.04456 \\
\bottomrule
\end{tabular}
\end{table}

\paragraph{Search trajectory contrast.}
The most striking difference between models is the exploration--exploitation balance.
GPT-5.2's Strategist maintained diversity throughout the campaign: its top-10 experiments span iTransformer, N-HiTS, TSMixer, TimesNet, PatchTST, and ridge stacking.
Opus's Strategist converged on TFT after experiment~${\sim}$10 and spent the remaining 40~experiments refining TFT hyperparameters (dropout, loss function, epochs, context length, ensemble methods).
As the playbook noted: \emph{``TFT is the clear winner -- Variable Selection Networks + attention + calendar known-future-inputs.''}
This focused exploitation happened to land on the best architecture (TFT at 0.02142 beat iTransformer at 0.02204), but the playbook also acknowledged the risk: 14~experiments (28\% of budget) were spent on post-processing methods that all failed to improve over the base TFT.

\paragraph{Opus's winning TFT configuration.}
Hidden dim 160, 8~attention heads, dropout 0.3, context 336h, sensor ID embedding (dim 48), MSE loss, cosine annealing LR (5e-4 to 1e-6), 80~epochs, batch size 64, 500~batches/epoch.
Total parameters: 2.1M.
Training time: 27~minutes on 1$\times$H100.

\subsection{Convergence curves}
\label{app:convergence}

\begin{figure}[h]
\centering
\includegraphics[width=\textwidth]{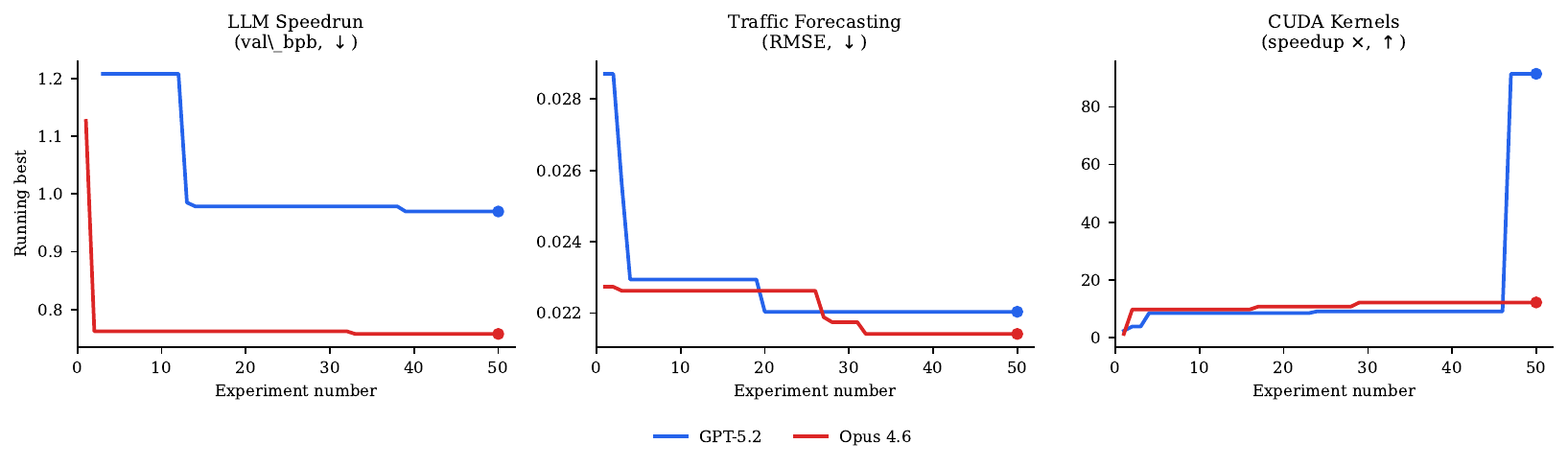}
\caption{Convergence curves across all three domains.  Each line shows the running-best metric (lower is better for LLM and traffic; higher for CUDA) as a function of analyzed experiment count.  Both models improve rapidly in the first 10--15 experiments and plateau by 25--30.}
\label{fig:convergence}
\end{figure}

Figure~\ref{fig:convergence} shows the running-best metric as a function of experiment number for all three domains and both models.

\paragraph{Convergence dynamics.}
All campaigns show a characteristic pattern: rapid improvement in experiments 1--15 as the system explores the most promising architectural families, followed by diminishing returns as it enters the refinement phase.
By experiment~25--30, the running-best metric has typically stabilized to within 5\% of the final value.

Convergence speed varies by domain.
CUDA kernels converge fastest because each experiment produces a clear, unambiguous speedup metric with low noise: the system can quickly identify which optimization strategies work (e.g., warp-shuffle reductions) and which don't (convolutions).
LLM pretraining converges more slowly due to higher variance: GPU contention introduces ${\sim}$0.05 BPB of noise, and architectural differences compound over 20-minute training runs.
Traffic forecasting falls in between, with moderate noise from the 7-window rolling evaluation.

The convergence criterion ($C{=}20$ experiments without improvement) was never triggered in our campaigns because we set a hard budget of 50~experiments.
In practice, most campaigns would have naturally terminated around experiment~35--40 based on the convergence curves.

\section{Playbook excerpts and dynamics}
\label{app:playbook}

\subsection{LLM speedrun playbook (GPT-5.2, after 50 experiments)}
\label{app:playbook-llm}

The GPT-5.2 LLM speedrun playbook evolved from a blank document to a comprehensive 2{,}000-word strategic reference over 50~experiments.
We highlight the most informative sections (verbatim, lightly formatted):

\paragraph{Verified SOTA entry.}
\begin{quote}\small\itshape
Verified SOTA: Exp \#14 -- \texttt{llama24\_512\_gqa\_swiglu\_adamw\_wsd}.
Best val\_bpb\_full20m\_best: 0.9786.  Params: 91.38M.
Arch: LLaMA-style (RMSNorm + RoPE + SwiGLU, pre-norm, tied embeddings).
Shape: 24 layers, d\_model=512, n\_head=8, GQA kv\_heads=2.
Optim: AdamW ($\beta$=(0.9,0.95), wd=0.1, clip=1.0) + WSD (lr=6e-4, warmup=800).
Data: byte tokenizer (vocab\_size=256), language\_filter=en, include\_query=true, shards 1--10, seq\_len=1024.
\end{quote}

\paragraph{Metric integrity warning.}
The Strategist discovered that the leaderboard was mixing smoke-test metrics (${\sim}$8.0 bpb, near-random) with full 20-minute metrics (${\sim}$0.98--1.21 bpb), inverting rankings:
\begin{quote}\small\itshape
Metric integrity warning (critical): The board leaderboard is currently mixing smoke/step-0 eval metrics and full 20-minute scored metrics.  This can invert rankings and lead to incorrect decisions.
\end{quote}

\paragraph{Infrastructure warnings.}
\begin{quote}\small\itshape
torch.compile has repeatedly failed with RoPE lazy cache init + Inductor/CUDAGraph overwrites.  Treat compile as opt-in only with cache pre-init and cudagraphs disabled.
\end{quote}

\paragraph{Post-budget triage.}
In the final experiments (\#43--\#50), the Strategist documented cascading infrastructure failures (CUDA OOM, IndentationError in the harness, SLURM job failures) and prioritized fix actions, demonstrating the playbook's role as an operational log as well as a strategic document.

\subsection{CUDA kernels playbook (GPT-5.2)}
\label{app:playbook-cuda}

The CUDA kernels playbook (GPT-5.2) developed a technique taxonomy and explicit avoidance rules.
Key excerpts:

\paragraph{Positive technique taxonomy.}
\begin{quote}\small\itshape
Diagonal matrix ops: $\text{diag}(A) \cdot B$ is just elementwise multiply of the diagonal vector with each row of $B$.  Yields 10--68$\times$ over PyTorch's full matmul.

Warp-shuffle reductions yield 73--75$\times$ on sum operations.  Use \texttt{\_\_shfl\_down\_sync} for warp-level parallel reduction; avoid shared memory round-trips.

Operator fusion (LayerNorm + residual, GEMM + LogSumExp): fuse sequential operations into a single kernel pass.  Use vectorized float4 loads/stores for memory bandwidth.  Yields 16--91$\times$.
\end{quote}

\paragraph{``Do not attempt'' list.}
\begin{quote}\small\itshape
Do not attempt convolution kernels -- cuDNN is too well-optimized; handwritten runs at 0.05--0.73$\times$.

CUTLASS-dependent kernels: the submodule is incomplete in this environment.  Do not attempt CUTLASS GEMM templates.

Bool-mask scatter operations: correctness failures due to non-contiguous memory layouts.  Avoid unless the operator is simple enough to verify by hand.
\end{quote}

\paragraph{Queue-pruning heuristics.}
The Strategist learned to cancel queued experiments proactively based on accumulating evidence:
\begin{quote}\small\itshape
If an optimization technique has failed on 2+ similar tasks, cancel all remaining variants.  Specifically: cancel all convolution experiments after L1\_t54 and L1\_t58 both showed 0.05--0.73$\times$.
\end{quote}

\subsection{Traffic forecasting playbook (Opus~4.6)}
\label{app:playbook-traffic}

The Opus traffic playbook demonstrates both the strengths and risks of playbook-driven convergence.

\paragraph{TFT convergence trajectory.}
The Strategist initially proposed diverse architectures (PatchTST, DeepAR, iTransformer, TFT, N-BEATS).
After experiment~${\sim}$10, TFT emerged as the clear leader (RMSE ${\sim}$0.0226 vs.\ ${\sim}$0.0227 for PatchTST).
The playbook recorded:
\begin{quote}\small\itshape
TFT is the clear winner -- Variable Selection Networks + attention + calendar known-future-inputs.
MSE loss directly optimizes RMSE; 3.9\% better than quantile loss.
Dropout 0.3 is critical; dropout=0.1 underperforms by ${\sim}$2\%.
8 attention heads $>$ 4 for this task.
\end{quote}

\noindent From experiment~15 onward, the Strategist devoted the remaining budget almost exclusively to TFT refinement and post-processing.

\paragraph{Hyperparameter findings.}
The playbook documented specific findings:
dropout 0.3 optimal (tested 0.1, 0.2, 0.3, 0.4);
MSE loss beats quantile for RMSE optimization;
80--150 epochs sufficient (diminishing returns beyond 100);
context 336h optimal (168h viable, 672h not consistently better).

\paragraph{The post-processing trap.}
The playbook's final campaign statistics reveal a cautionary tale: 14~experiments (28\% of budget) were spent on post-processing methods (bias correction, residual MLP, per-horizon calibration, ensemble methods), all of which failed to meaningfully improve over the best single TFT.
The playbook explicitly noted: \emph{``TFT residuals are near-random (SNR ${\sim}$0.04) -- post-processing cannot extract signal from noise.''}

\paragraph{iTransformer absence.}
Opus never explored iTransformer -- the architecture that GPT-5.2 found to be the strongest.
This represents the fundamental tension in playbook-driven search: early convergence on TFT was \emph{correct} in that TFT turned out to be the best architecture Opus found, but it may have been \emph{suboptimal} in that broader exploration might have discovered the iTransformer + per-horizon calibration trick that GPT-5.2 found (RMSE 0.01837 in the full evaluation, substantially better than Opus's 0.02142).

\subsection{Playbook growth dynamics}
\label{app:playbook-dynamics}

We analyze playbook dynamics across domains and models.

\paragraph{Growth trajectory.}
Playbooks grow rapidly in the first 15~experiments as the Strategist documents initial findings, then transition to consolidation and refinement.
The Opus LLM speedrun playbook reached ${\sim}$2{,}500 words by experiment~50, organized into 7~major sections (objective, definitive findings, tricks verdict, confirmed losers, GPU contention analysis, best configuration reference, and lessons for future work).
The GPT-5.2 traffic playbook reached ${\sim}$800 words by experiment~50, more operational in tone (priority reruns, evaluation integrity gates, shipping rules).

\paragraph{Self-correction events.}
We observed several instances of the Strategist invalidating its own prior playbook entries:
\begin{itemize}[nosep]
  \item \textbf{LLM speedrun (Opus)}: Early playbook entries recommended deeper architectures (20L$\times$576d). After experiments~15--20 showed these underperforming, the playbook was updated: \emph{``$\geq$14 layers: Too deep for 20-minute budget''} and \emph{``Muon optimizer: Always 5--12\% worse than AdamW (6 head-to-head tests).''}
  \item \textbf{Traffic (Opus)}: Early enthusiasm for post-processing methods was reversed after 14~failed attempts: \emph{``ALL post-processing/bias correction at full scale failed -- TFT residuals are near-random.''}
\end{itemize}

\paragraph{Cross-model comparison.}
Opus playbooks tend to be more structured and analytical (explicit rankings, comparative tables, quantified findings), while GPT-5.2 playbooks are more operational (action items, priority lists, fix instructions).
Opus's traffic playbook contains a definitive tier list (S/A/B/C), campaign statistics, and unrealized ideas for future work; GPT-5.2's traffic playbook focuses on evaluation integrity, root cause analysis of failures, and priority reruns.
Both styles are effective: Opus's analytical approach enabled tight convergence on TFT, while GPT-5.2's operational approach helped navigate infrastructure issues (missing data paths, SLURM failures).

\section{User interface}
\label{app:ui}

\system includes a web-based dashboard (React/TypeScript frontend, FastAPI backend with WebSocket streaming) for real-time campaign monitoring.
The interface is organized into three resizable panes:

\paragraph{Kanban board.}
The central view displays experiments organized by status across 9~columns: \texttt{to\_implement}, \texttt{implemented}, \texttt{checked}, \texttt{queued}, \texttt{running}, \texttt{finished}, \texttt{analyzed}, \texttt{done}, and \texttt{cancelled}.
Each experiment is represented as a card showing its name, hypothesis, current metric (if available), assigned Worker ID, and SLURM job ID.
Columns are collapsible and support pagination for large campaigns.

\paragraph{Leaderboard.}
A sortable, filterable table ranking all analyzed experiments by the primary metric (configurable per domain).
The leaderboard auto-refreshes every 10~seconds and on new experiment events, allowing the human to track progress in real time.

\paragraph{File tree and viewer.}
The left pane provides a hierarchical file browser of the workspace, with syntax-highlighted code viewing for generated scripts, model implementations, and configuration files.

\paragraph{Conversation stream.}
The right pane shows a real-time log of agent tool calls and results (shell commands executed, files read, web searches performed), providing full transparency into the system's reasoning and actions.

\paragraph{Human-in-the-loop interface.}
A chat panel allows the human operator to ask questions during a campaign (e.g., ``What's happening?'', ``Any errors?'', ``Best model so far?'').
We are experimenting with allowing the human to suggest specific experiments, veto proposals, or add notes to the playbook.
Early observations suggest that even minimal human guidance (e.g., ``focus on TFT variants'' or ``stop trying convolution kernels'') can significantly accelerate convergence, though quantifying this effect requires controlled experiments we have not yet run.

\paragraph{Milestone reports.}
Periodic summaries generated by the Reporter agent are displayed in the status view, showing cumulative progress, flagged experiments (suspicious metrics, smoke-test-only results), and recommendations for the Strategist.

\begin{figure}[h]
\centering
\includegraphics[width=\textwidth]{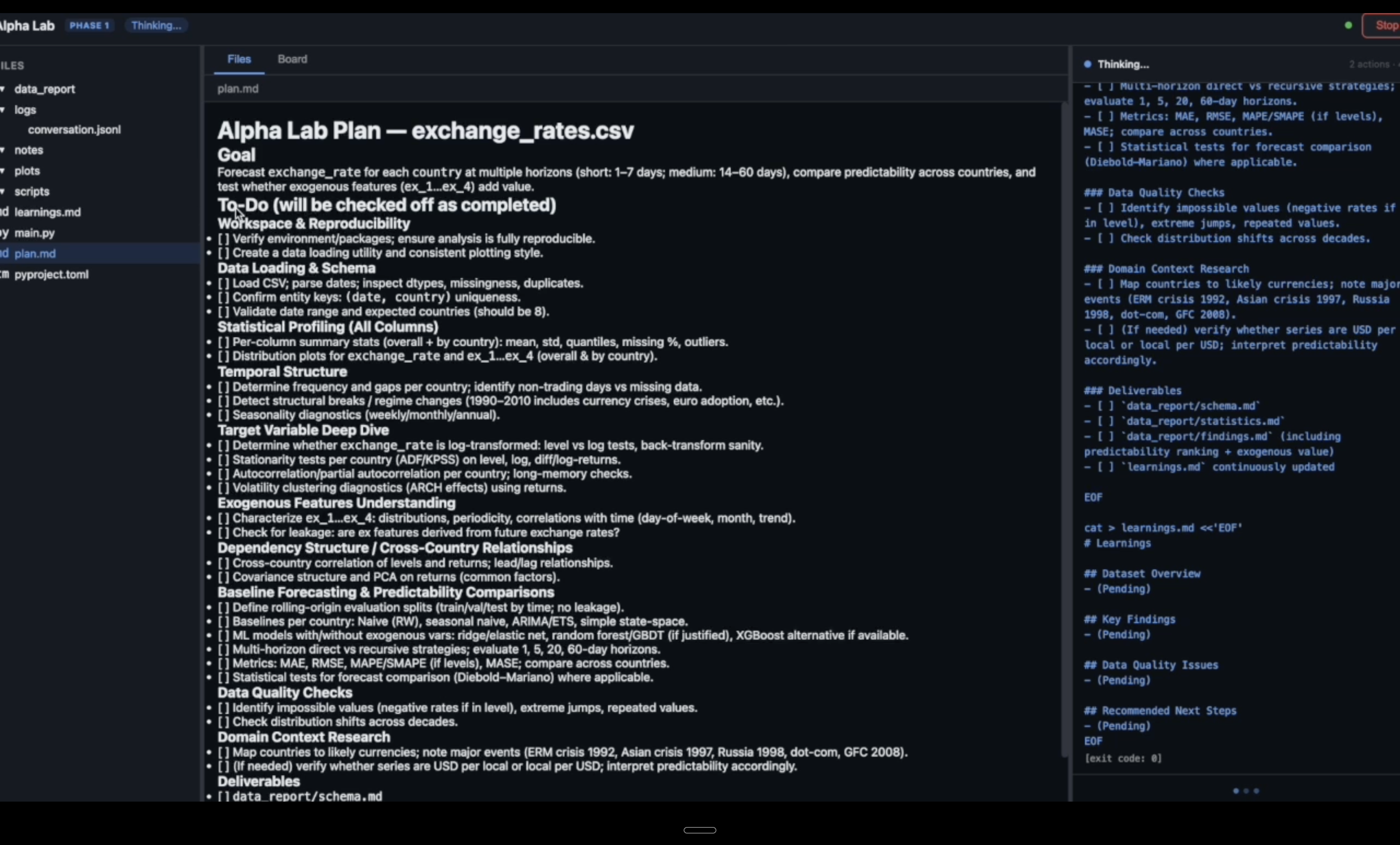}
\caption{\textbf{Phase~1 dashboard: Explorer agent plan.}  The Explorer agent has autonomously generated a detailed to-do list (\texttt{plan.md}, center pane) for the exchange-rate forecasting domain, covering data loading, temporal structure analysis, stationarity tests, cross-country dependencies, baseline comparisons, and domain context research.  Items are checked off as the agent completes them.  The left pane shows the workspace file tree (scripts, notes, plots being generated); the right pane shows the agent's real-time thinking and tool calls.}
\label{fig:ui-phase1-todo}
\end{figure}

\begin{figure}[h]
\centering
\includegraphics[width=\textwidth]{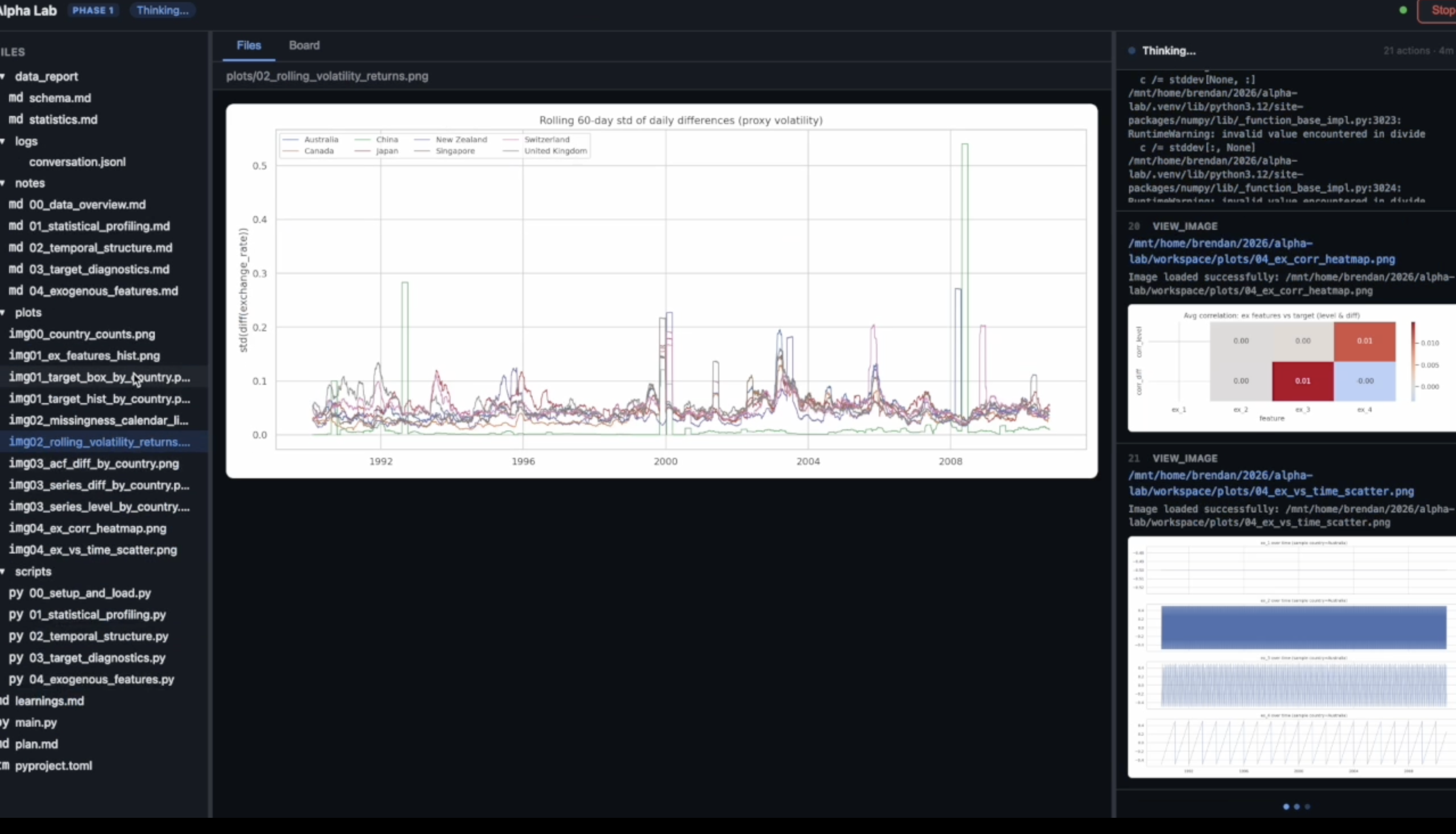}
\caption{\textbf{Phase~1 dashboard: data exploration in progress.}  The Explorer agent has generated analytical plots (center pane shows a rolling volatility time series across currency pairs) and is viewing them via the \texttt{view\_image} tool (right pane, with additional correlation heatmaps and scatter plots).  The left pane shows the growing collection of analysis scripts and generated plots in the workspace.}
\label{fig:ui-phase1-plots}
\end{figure}

\begin{figure}[h]
\centering
\includegraphics[width=\textwidth]{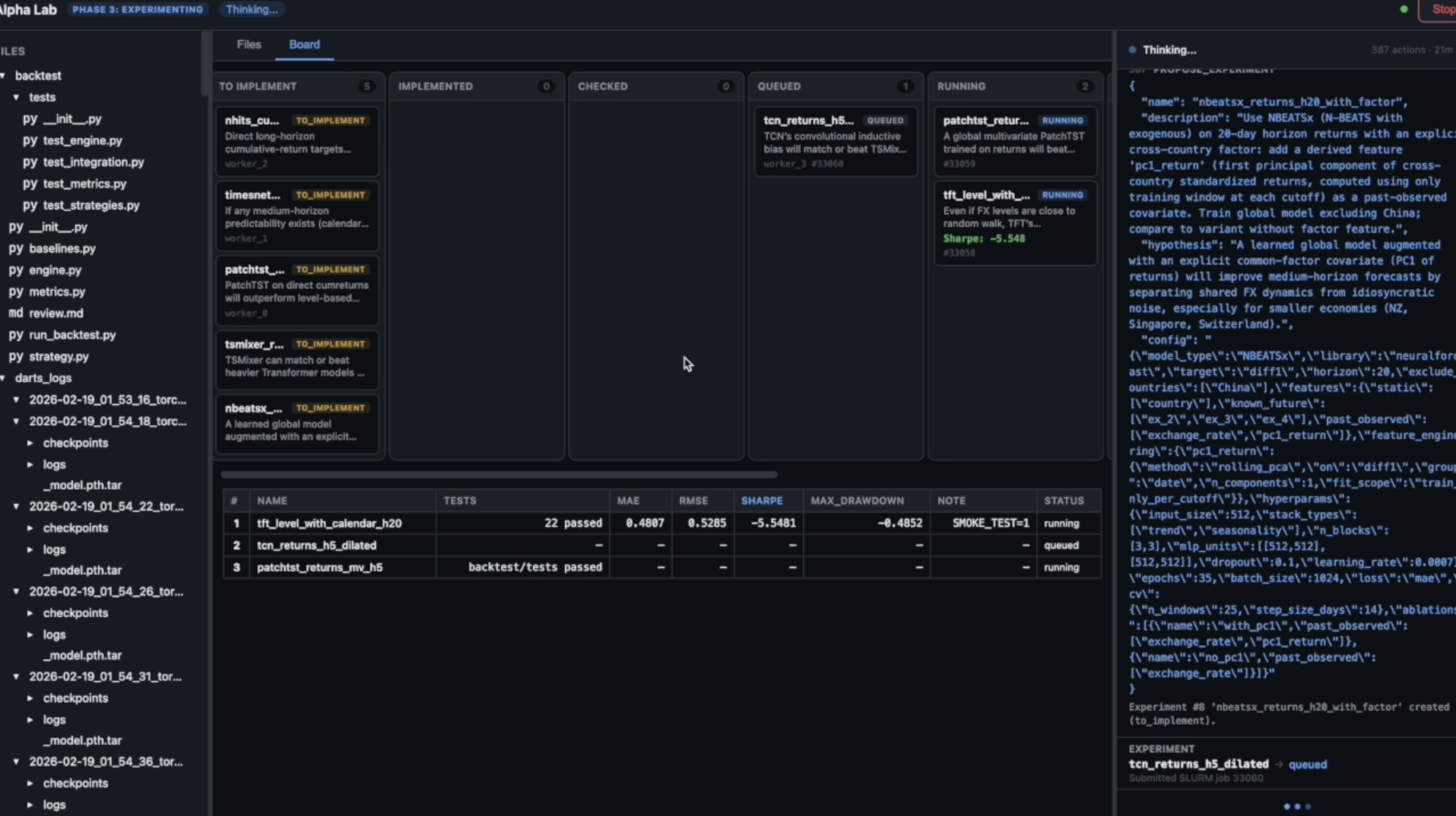}
\caption{\textbf{Phase~3 dashboard: GPU experimentation.}  The Kanban board (center, top) shows experiments in various stages: \texttt{to\_implement}, \texttt{implemented}, \texttt{checked}, \texttt{queued}, and \texttt{running}.  Each card displays the experiment name, hypothesis, and current status.  The leaderboard (center, bottom) ranks completed experiments by the primary metric.  The left pane shows the workspace file tree with experiment directories, checkpoints, and logs.  The right pane shows the conversation stream of the currently active Worker agent implementing an experiment.}
\label{fig:ui-phase3}
\end{figure}

\section{Additional experiments}
\label{app:additional}

\subsection{Financial time series forecasting (exchange rates)}
\label{app:exchange}

This domain tests \system on a task with no built-in adapter; Phase~0 must generate everything from scratch.

\paragraph{Data.}
Eight synthetic daily exchange-rate series against the US dollar, spanning 6{,}071 business days (1990--2013).
The dataset is stored in GluonTS format with a 30-business-day prediction horizon and 5 rolling-origin test windows per currency (40 test records total).
Values range from ${\sim}$0.006 (item~4, a peg-like currency) to ${\sim}$2.1, with substantial cross-series heterogeneity.

\paragraph{Task.}
Forecast 30 business days ahead; evaluate via annualized Sharpe ratio (primary, maximize), RMSE, max drawdown, and directional accuracy.
The trading evaluation maps forecasts to positions via $\text{position} = \text{sign}(\text{predicted 30-day return})$, then computes PnL from realized returns.

\paragraph{Phase~0: adapter generation.}
Since ``exchange rates'' does not match any built-in adapter, the Phase~0 generation agent was invoked.
It examined the data (detecting the panel structure, running ADF stationarity tests), searched the web for FX forecasting best practices, and generated all 11~adapter files from scratch.
Key design decisions made by the agent: use log-returns rather than levels, walk-forward split with embargo periods to prevent lookahead, Sharpe ratio as the primary metric, and max drawdown as a secondary safety metric.
The generated \texttt{domain\_knowledge.md} included guidance on per-series normalization (fit on train window only to avoid leakage), global models with item embeddings to share strength across 8~series, and probabilistic forecasting given FX noisiness.

\paragraph{Phase~1: data exploration.}
The Explorer autonomously generated a 9-section plan and executed it over several hours, producing 9~analysis scripts, 12~plots, and 11~detailed notes.
The completed plan (all items checked off by the agent):

\begin{quote}\small\ttfamily
\begin{enumerate}[nosep,leftmargin=*]
\item[\checkmark] Setup \& Intake: locate dataset, validate 8 series, build DataFrames
\item[\checkmark] Schema \& Profiling: per-series stats, data quality checks
\item[\checkmark] Temporal Structure: date ranges, frequency validation, regime detection
\item[\checkmark] Return Transformations: levels vs returns, ADF/KPSS stationarity tests
\item[\checkmark] Cross-Series Dependency: correlation, PCA, lead-lag analysis
\item[\checkmark] Baselines: naive, drift, AR(1), ARIMA, VAR on returns
\item[\checkmark] Trading Backtests: sign-of-return strategies, transaction cost sensitivity
\item[\checkmark] Model Recommendations: classical + DL families
\item[\checkmark] Final Deliverables: schema.md, statistics.md, findings.md
\end{enumerate}
\end{quote}

Key findings recorded in \texttt{learnings.md}:
\begin{itemize}[nosep]
  \item FX levels are non-stationary (near random walks); log-returns are approximately stationary with heavy tails and strong volatility clustering.
  \item Return autocorrelation is near zero -- weak mean predictability, but volatility/risk scaling is more promising.
  \item Strong cross-series correlations (max 0.80 between items~0 and~6); PCA shows 45\% of variance in PC1 (common factor), 19\% in PC2.
  \item Item~4 exhibits peg-like behavior with step changes, warranting separate handling.
  \item Rolling-origin baseline trading yields Newey--West adjusted Sharpe ${\approx}$0.16 (univariate drift/AR(1)); multivariate VAR(1) improves marginally to ${\approx}$0.17.
\end{itemize}

\begin{figure}[h]
\centering
\begin{subfigure}[b]{0.48\textwidth}
\includegraphics[width=\textwidth]{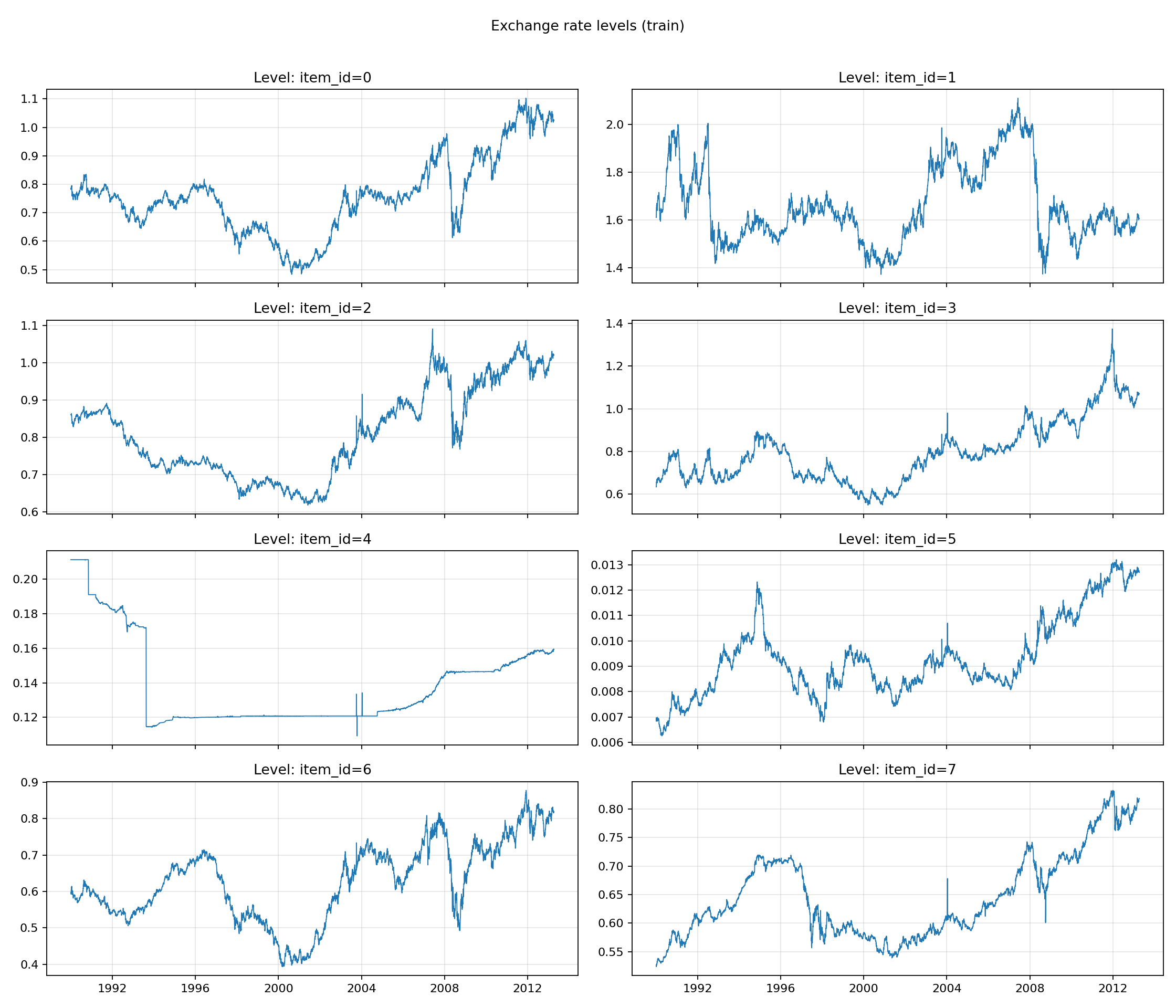}
\caption{FX levels by currency (8 series, 1990--2013).}
\end{subfigure}
\hfill
\begin{subfigure}[b]{0.48\textwidth}
\includegraphics[width=\textwidth]{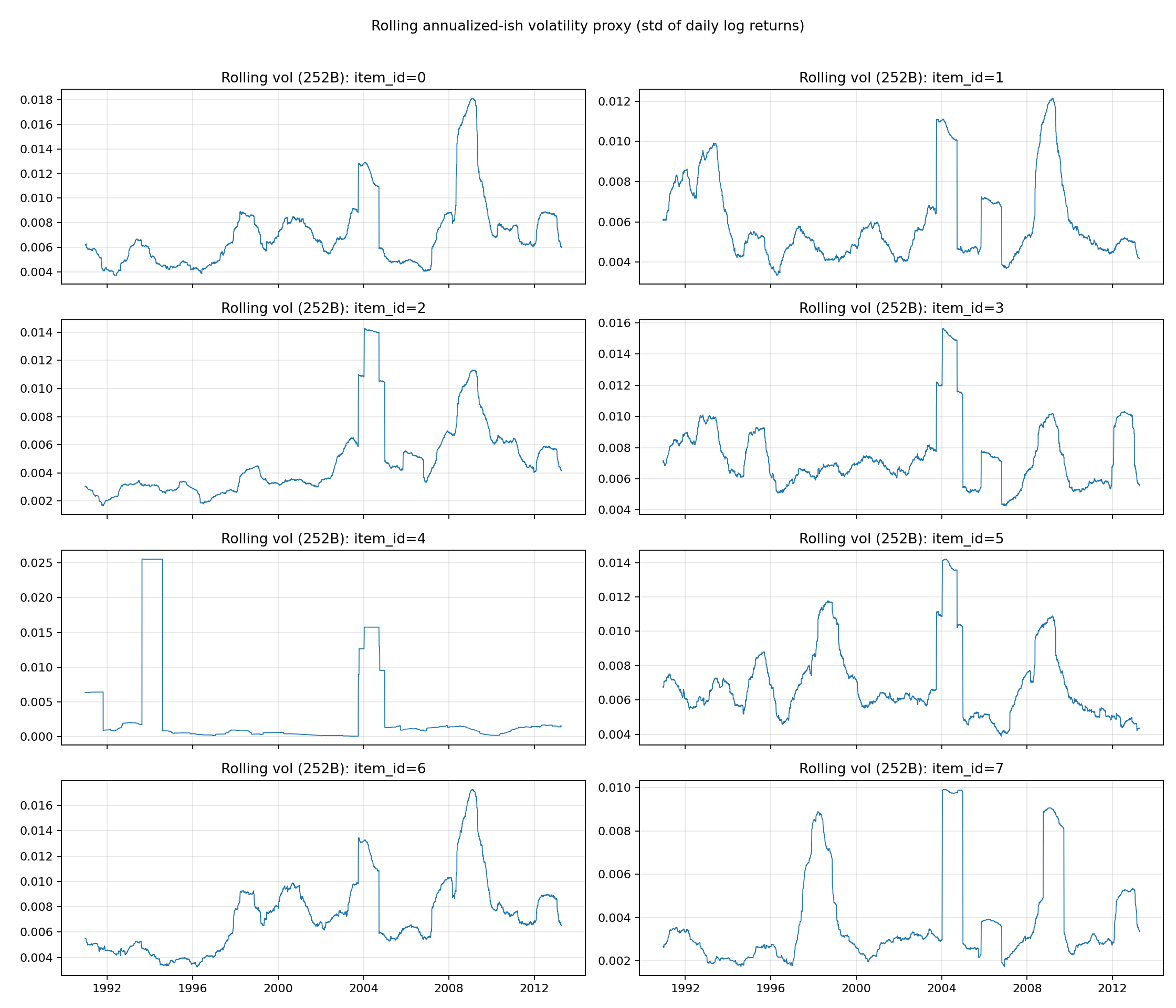}
\caption{Rolling 252-day volatility, showing regime shifts (${\sim}$1997, 2004, 2008).}
\end{subfigure}
\\[6pt]
\begin{subfigure}[b]{0.48\textwidth}
\includegraphics[width=\textwidth]{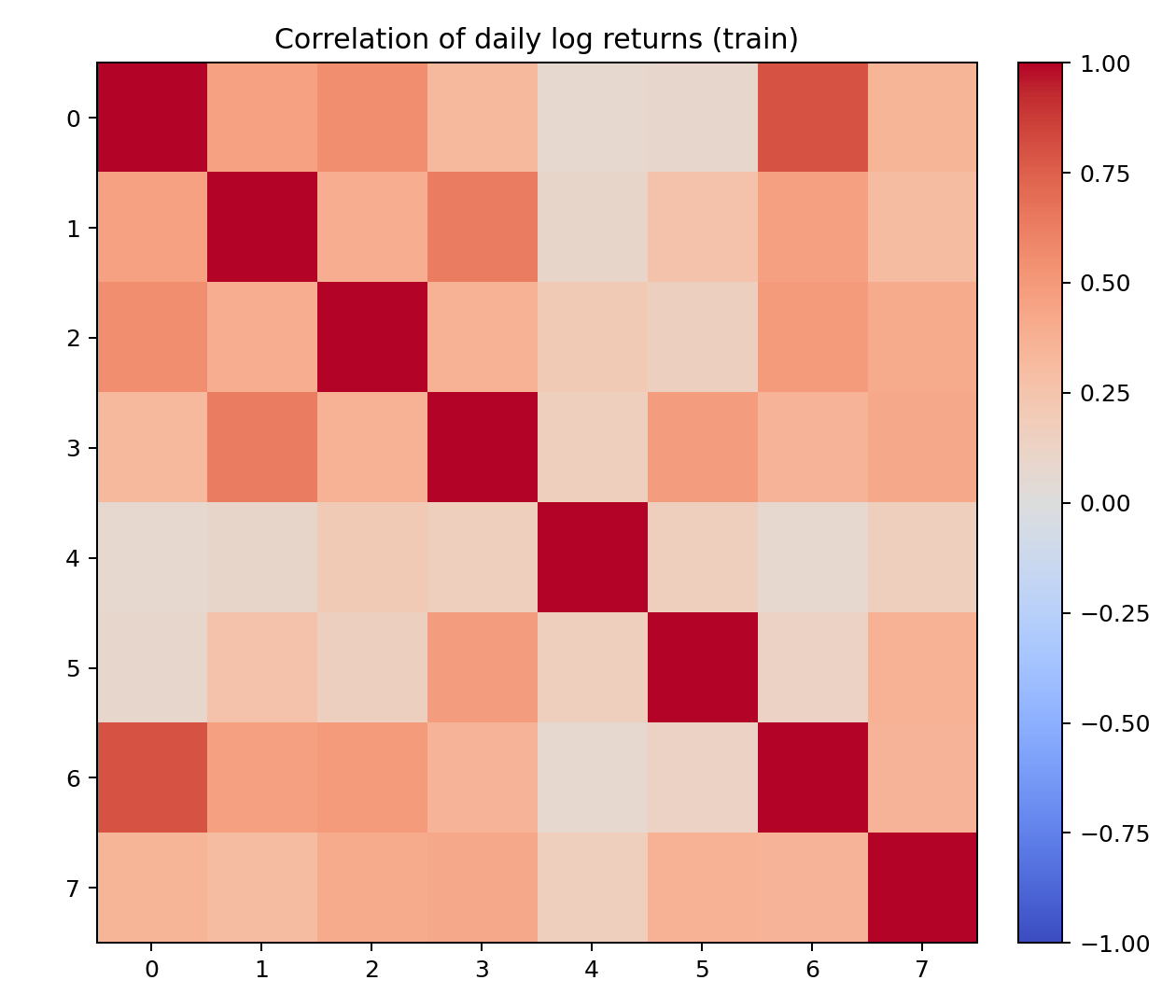}
\caption{Return correlation heatmap.}
\end{subfigure}
\hfill
\begin{subfigure}[b]{0.48\textwidth}
\includegraphics[width=\textwidth]{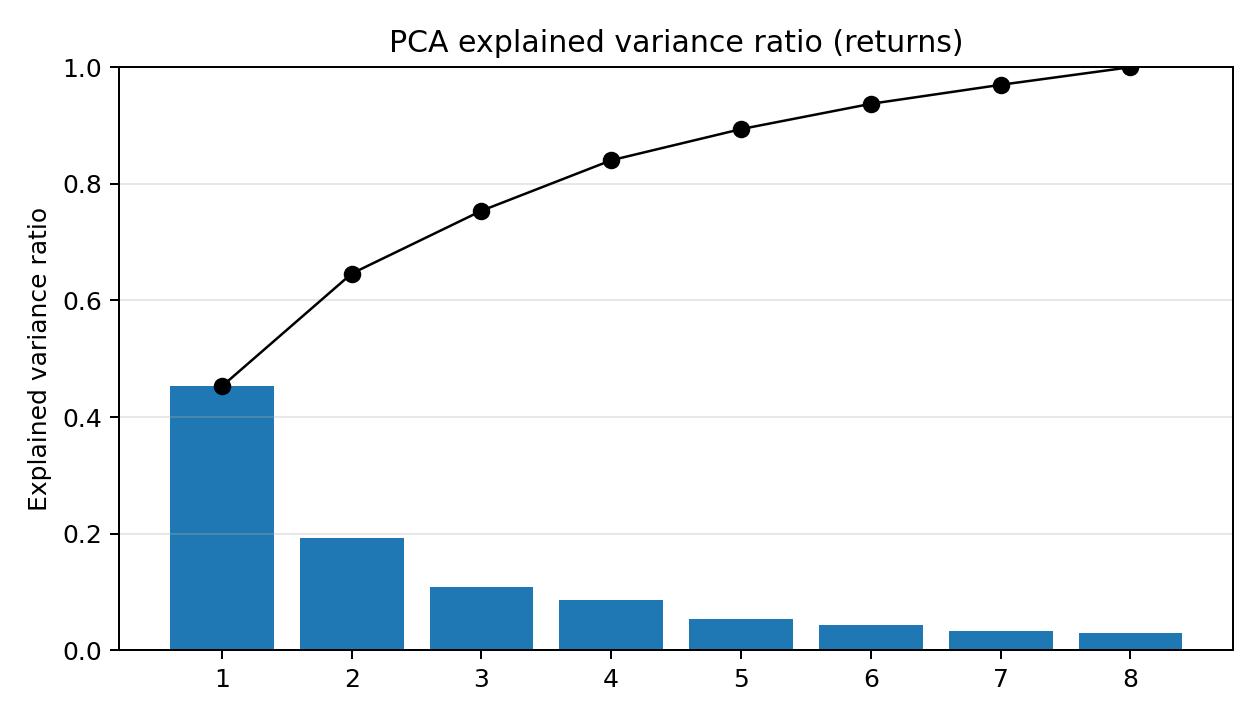}
\caption{PCA explained variance (PC1 captures 45\%).}
\end{subfigure}
\caption{Phase~1 exploration plots generated autonomously by \system for the exchange-rate domain.  The Explorer wrote the analysis scripts, generated these visualizations, and used the \texttt{view\_image} tool to inspect them before recording findings.}
\label{fig:fx-exploration}
\end{figure}

\paragraph{Phase~2: evaluation harness.}
The Builder produced a walk-forward backtesting framework with a \texttt{Strategy} abstract base class (requiring \texttt{fit(y\_train)} and \texttt{predict(y\_context)} methods), a \texttt{WalkForwardEngine} with configurable embargo, and metrics including Sharpe, max drawdown, directional accuracy, and RMSE.
The Critic caught a subtle lookahead bug: a rolling volatility computation in the feature engineering pipeline used the current day's close price, which would not be available at prediction time.
The Tester validated the fix, and the loop converged in 3~iterations.
The Critic's review also flagged that the embargo parameter only affects \texttt{fit()} but not \texttt{predict()}'s context window -- a non-blocking issue that was documented for future improvement.

\paragraph{Phase~3: GPU-scale experimentation.}
The system ran 43~experiments (29 with valid Sharpe ratios) spanning a wide range of architectures: PatchTST, TimesNet, DeepAR, DeepVAR, TFT, N-HiTS, N-BEATSx, TCN, TSMixer, Mamba SSM, TS2Vec, and a custom Transformer policy network trained end-to-end on a differentiable Sharpe loss.
Table~\ref{tab:fx-leaderboard} shows the full leaderboard.

\begin{table}[h]
\centering
\caption{Exchange-rate experiment leaderboard (annualized Sharpe, higher is better).  All experiments use GPT-5.2.  Sharpe values are computed over only 5 rolling-origin windows per currency (40 trades total), making individual estimates high-variance.  $\dagger$~denotes experiments flagged by the system's own debrief as potentially unreliable.}
\label{tab:fx-leaderboard}
\small
\begin{tabular}{@{}rlrl@{}}
\toprule
\textbf{Rank} & \textbf{Sharpe} & \textbf{RMSE} & \textbf{Experiment} \\
\midrule
1$^\dagger$  & 4.214  & 0.0121 & TimesNet multivariate Student-T \\
2$^\dagger$  & 2.780  & 0.0141 & PatchTST factor-first PCA(3) + reconstruct \\
3            & 0.748  & 0.0163 & Transformer policy net (Sharpe loss) \\
4            & 0.697  & 0.0185 & Stacking ensemble (turnover-penalized) \\
5            & 0.395  & 0.0125 & TCN multivariate risk-parity \\
6            & 0.121  & ---    & TSMixer global returns \\
7            & 0.077  & 0.0160 & TFT cross-currency with lag covariates \\
\midrule
8--29        & $-$0.05 to $-$23.9 & 0.011--0.022 & (22 experiments with negative Sharpe) \\
\midrule
\multicolumn{2}{l}{Baseline: drift} & --- & NW-Sharpe $\approx$ 0.16 \\
\multicolumn{2}{l}{Baseline: VAR(1)} & --- & NW-Sharpe $\approx$ 0.17 \\
\bottomrule
\end{tabular}
\end{table}

The top two results (TimesNet at 4.21 and PatchTST factor-first at 2.78) were flagged by the system's own debriefs as unreliable: both are computed from only 5 non-overlapping 30-day trades per currency, leverage was 5--8$\times$, and no Newey--West correction was applied.
The system explicitly noted: \emph{``Treat as high-variance, not reliable evidence of edge until gating passes.''}

The most interesting experiment was the Transformer policy network (rank~3, Sharpe 0.748), which trained end-to-end on a differentiable Sharpe-like surrogate loss:
\begin{quote}\small\itshape
Input: 512-day context of $[r_t, \text{EWMA\_vol}_{10}, \text{EWMA\_vol}_{30}, \text{EWMA\_vol}_{90}]$.
Model: 4-layer Transformer encoder, 8 heads, $d_\text{model}{=}128$, per-currency item embeddings.
Loss: $-\text{mean}(\text{PnL})/\text{std}(\text{PnL})$ with leverage and turnover penalties.
\end{quote}
The system's debrief identified five specific limitations: only 40 trades total; no Newey--West adjustment despite overlapping horizons; the evaluation discards position magnitude (reducing the learned continuous policy to sign); models are fit per-currency despite item embeddings; and zero transaction costs.
This level of self-critical analysis (automatically generated by the Worker agent during the \texttt{analyze} phase) is representative of the quality of debriefs across domains.

\begin{figure}[h]
\centering
\begin{subfigure}[b]{0.48\textwidth}
\includegraphics[width=\textwidth]{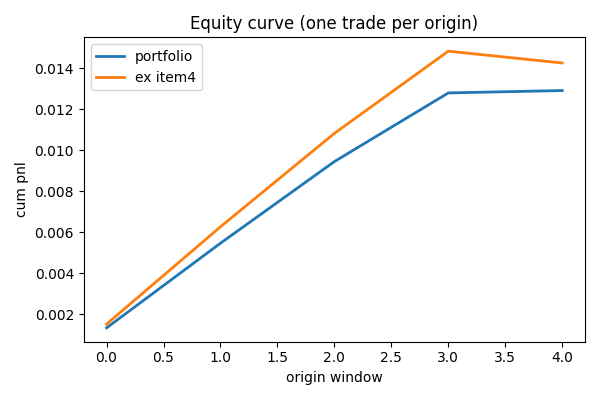}
\caption{TimesNet multivariate (Sharpe 4.21$^\dagger$).}
\end{subfigure}
\hfill
\begin{subfigure}[b]{0.48\textwidth}
\includegraphics[width=\textwidth]{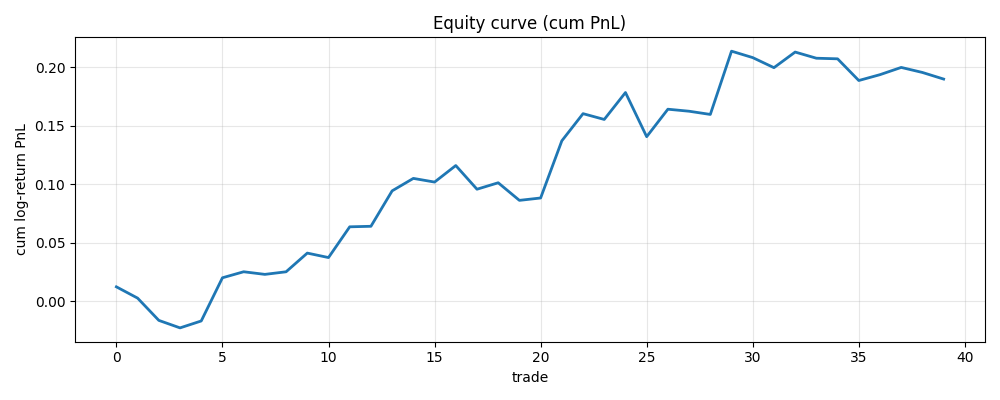}
\caption{Transformer policy net (Sharpe 0.75).}
\end{subfigure}
\caption{Equity curves for the top-ranked and most interesting exchange-rate experiments, generated by \system.  The TimesNet result (left) exhibits suspiciously smooth performance from only 5 trades per currency; the Transformer policy net (right) shows more realistic variance.}
\label{fig:fx-equity}
\end{figure}

\paragraph{Playbook evolution.}
The playbook accumulated several domain-specific lessons over the campaign:
\begin{itemize}[nosep]
  \item \emph{``Model log-returns, not levels -- ADF tests confirm levels are non-stationary.''}
  \item \emph{``Per-series normalization must be fit on train window only; global normalization causes leakage.''}
  \item \emph{``Multivariate training on returns is beneficial vs univariate -- cross-currency correlations (max 0.80) provide real signal.''}
  \item \emph{``Sharpe estimates from 5 trades are extremely high-variance; do not trust absolute values.''}
  \item \emph{``Exclude item\_id=4 from pooled metrics or model separately (peg-like with step changes).''}
\end{itemize}

\paragraph{Discussion.}
This domain demonstrates \system's ability to handle genuinely novel tasks: the auto-generated adapter was functional and reasonable, correctly identifying the key design decisions (log-returns, walk-forward, Sharpe metric) without human guidance.
The system's self-critical debriefs (flagging unreliable results, identifying statistical limitations, and suggesting follow-up experiments) show that the analyze phase produces genuine research insight, not just metric extraction.
The quality gap between the auto-generated adapter and hand-written built-in adapters was modest, mainly in prompt specificity: the generated prompts were more generic, lacking the domain-specific guardrails (e.g., ``never set deterministic algorithms on H100'') that built-in adapters accumulate over time.
The fact that 22 of 29 experiments produced negative Sharpe ratios also illustrates the difficulty of the domain (FX returns are notoriously hard to predict) and the system's willingness to report honest failures rather than cherry-pick results.

\section{Cost and token breakdown}
\label{app:cost}

We extracted token usage from the JSONL conversation logs recorded during each campaign.
Table~\ref{tab:token-usage} summarizes the token consumption across all primary runs.
All cost figures in this section reflect \textbf{LLM API token costs only} and do not include GPU compute.
Our experiments ran on on-premise 4$\times$H100 hardware; at current cloud rates (${\sim}$\$2--3/GPU-hour for H100 instances), the 12--48 hours of GPU time per campaign would add ${\sim}$\$100--600 depending on domain and utilization.

\begin{table}[h]
\centering
\caption{Token usage and API call counts for primary campaigns.  Output tokens are consistently ${\sim}$1--2\% of input tokens.}
\label{tab:token-usage}
\small
\begin{tabular}{@{}llrrrr@{}}
\toprule
\textbf{Domain} & \textbf{Model} & \textbf{Input tokens} & \textbf{Output tokens} & \textbf{API calls} & \textbf{Est.\ cost} \\
\midrule
LLM Speedrun & GPT-5.2 & 84.3M & 979K & 2{,}892 & $\sim$\$150 \\
LLM Speedrun & Opus~4.6 & 116.9M & 1.5M & 2{,}512 & $\sim$\$200 \\
LLM Speedrun & Sonnet~4.6 & 195.3M & 3.4M & 4{,}192 & $\sim$\$120 \\
Traffic & GPT-5.2 & 73.5M & 1.4M & 3{,}415 & $\sim$\$180 \\
Traffic & Opus~4.6 & 92.2M & 1.7M & 2{,}787 & $\sim$\$200 \\
Electricity & GPT-5.2 & 133.8M & 1.8M & 4{,}963 & $\sim$\$190 \\
Electricity & Opus~4.6 & 144.7M & 1.8M & 4{,}289 & $\sim$\$200 \\
\bottomrule
\end{tabular}
\end{table}

\paragraph{Token distribution by phase.}
Phase~3 dominates token consumption, accounting for ${\sim}$97--98\% of total tokens across all campaigns.
This is expected: Phase~3 involves dozens of Strategist turns and Worker implementation/analysis/fix cycles, each requiring substantial context (leaderboard, playbook, debriefs, code).
Phase~1 accounts for ${\sim}$1--2\%, Phase~2 for ${\sim}$0.5--1\%, Phase~0 for $<$0.1\%, and Supervisor interventions for ${\sim}$0.3--0.5\%.

\paragraph{Cost per experiment.}
For a typical 50-experiment campaign, the average cost per experiment is \$3--4.
However, this varies substantially: early experiments (where the Strategist has limited context) cost less than late experiments (where the playbook, leaderboard, and debrief history are large).
The Strategist's context window grows roughly linearly with campaign length, driving the late-campaign cost increase.

\paragraph{Model cost comparison.}
Opus campaigns cost ${\sim}$\$200 (higher per-token pricing), GPT-5.2 campaigns cost ${\sim}$\$150--190 (lower per-token but sometimes more total tokens due to longer outputs), and Sonnet campaigns cost ${\sim}$\$120 (lower per-token pricing despite high token counts, due to Sonnet's verbose generation style resulting in more API calls).

\paragraph{Variance runs.}
The 5~GPT-5.2 variance runs (Phase~3 only, shared Phase~1/2) consumed 74--97M input tokens each (${\sim}$\$50--70 per run), confirming that Phase~3 alone accounts for the vast majority of campaign cost.
The total cost across all experimental runs (including variance and ablation runs) was approximately \$2{,}500.

\section{Failure analysis}
\label{app:failures}

We categorize failures observed across all campaigns into three tiers of severity.

\paragraph{Programmatic errors (Tier 1: self-healing).}
Runtime failures in LLM-generated code are the most common failure mode.
Examples include: CUDA out-of-memory errors (batch size too large for H100 VRAM), PyTorch API breaking changes (PyTorch 2.9.1 removed \texttt{.total\_mem} and \texttt{ReduceLROnPlateau(verbose=)}), Python import errors (package version mismatches), and SLURM job configuration issues (missing environment variables, incorrect paths).

Failure rates vary substantially by domain and model:
\begin{itemize}[nosep]
  \item \textbf{LLM speedrun}: GPT-5.2 had a 38\% failure rate (17/45 experiments produced no valid metrics), largely due to \texttt{torch.compile} failures and harness bugs.  Opus had only a 3\% failure rate (1/37).
  \item \textbf{Traffic}: Both models had low failure rates ($<$5\%), as the forecasting models are simpler and less prone to CUDA errors.
  \item \textbf{Electricity}: GPT-5.2 had a 55\% failure rate, the highest across all campaigns, driven by PyTorch 2.9.1 API changes affecting multiple deep learning libraries.
  \item \textbf{CUDA kernels}: 9/103 incorrect for GPT-5.2 (8.7\%) and 8/59 for Opus (13.6\%), primarily correctness failures rather than runtime crashes.
\end{itemize}

These errors are generally self-healing through two mechanisms: the Worker fix cycle (up to $k{=}2$ repair attempts per experiment) and the Supervisor's health check (which patches \texttt{domain\_knowledge.md} when error rates exceed $\tau{=}0.4$).

\paragraph{Evaluation errors (Tier 2: silent corruption).}
Bugs in the evaluation framework that slip past the Phase~2 Critic/Tester loop.
These are rarer but more dangerous because they silently corrupt all downstream results.
Examples caught during manual review:
\begin{itemize}[nosep]
  \item \textbf{Electricity}: The GPT-5.2 campaign produced several suspiciously low RMSE values (21.82 for an ensemble, 89.19 for LightGBM), likely due to data leakage in feature engineering.  These were flagged in the milestone reports but not automatically excluded.
  \item \textbf{CUDA}: The Critic caught a \texttt{torch.allclose} bug where the optimized output was passed as the \emph{reference} argument, making correctness checks trivially permissive.  This was fixed before Phase~3.
  \item \textbf{Traffic}: The GPT-5.2 playbook noted that some experiments reported metrics from smoke tests (10~instances) rather than full evaluation (6034~instances), polluting the leaderboard.
\end{itemize}

\paragraph{Strategic failures (Tier 3: going off the rails).}
Cases where the Strategist enters an unproductive loop or the playbook accumulates incorrect knowledge.
These are the hardest to detect and the most costly.
\begin{itemize}[nosep]
  \item \textbf{Premature convergence}: Opus on traffic locked onto TFT after experiment~${\sim}$10, spending 40~experiments on TFT refinement and never exploring iTransformer (which GPT-5.2 found to be stronger).  This \emph{happened} to produce the best single-model result, but the opportunity cost is unknown.
  \item \textbf{Scattered search}: GPT-5.2's variance runs v2 and v3 on LLM speedrun never converged on a productive architectural region, producing best val\_bpb of 1.011 and 1.006 respectively (vs.\ 0.964 for v1 and 0.970 for the primary run).
  \item \textbf{Budget waste on post-processing}: Opus's traffic campaign spent 14/50 experiments (28\% of budget) on post-processing methods that all failed to improve over the base TFT.
\end{itemize}

We estimate that approximately 1~in~5 campaigns would benefit from human intervention to redirect the Strategist.
The dashboard (Appendix~\ref{app:ui}) is designed to make such interventions lightweight: the human can monitor the playbook evolution, spot premature convergence, and inject suggestions via the chat panel.
The right balance between autonomy and oversight is an open research question; our experience suggests that a brief human check every 15--20 experiments is sufficient to catch strategic failures while preserving the system's autonomous operation.

\section{Reproducibility details}
\label{app:repro}

Full code, configuration files, adapter templates, and instructions for reproducing all experiments will be released on GitHub at \url{https://brendanhogan.github.io/alphalab-paper/}.

\paragraph{Model versions.}
All models were accessed in February--March 2026.
\begin{itemize}[nosep]
  \item \textbf{GPT-5.2}: Model ID \texttt{gpt-5.2}, accessed via OpenAI Responses API.  Reasoning effort set to \texttt{low} for most campaigns.
  \item \textbf{Claude Opus~4.6}: Model ID \texttt{claude-opus-4-6-v1}, accessed via AWS Bedrock Converse API.
  \item \textbf{Claude Sonnet~4.6}: Model ID \texttt{claude-sonnet-4-6-v1}, accessed via AWS Bedrock Converse API.
  \item \textbf{GPT-5.1-mini}: Model ID \texttt{gpt-5.1-mini}, tested on LLM speedrun but failed to produce trustworthy results: the model could not correctly implement the nats-to-BPB conversion, could not reliably diagnose environment errors, and exhausted its 50-experiment budget on runs with broken evaluation. The Strategist self-diagnosed the problem in the playbook but only after the budget was spent.
\end{itemize}

\paragraph{API parameters.}
All default temperature and sampling settings were used per provider.

\paragraph{Hardware.}
All experiments ran on a single node with 4$\times$ NVIDIA H100 NVL 80\,GB GPUs (sm\_90, 3.35\,TB/s HBM3 bandwidth), CUDA 12.6, PyTorch 2.9.1+cu126.
Time limits: 1{,}200\,s (20 minutes) for LLM speedrun, 7{,}200\,s (2 hours) for traffic and CUDA kernels.

\end{document}